\documentclass[10pt,twocolumn,letterpaper]{article}

\usepackage{iccv}
\usepackage{times}
\usepackage{epsfig}
\usepackage{graphicx}
\usepackage{amsmath}
\usepackage{amssymb}

\usepackage{listings}
\usepackage{xcolor}

\definecolor{codegreen}{rgb}{0,0.6,0}
\definecolor{codegray}{rgb}{0.5,0.5,0.5}
\definecolor{codepurple}{rgb}{0.58,0,0.82}
\definecolor{backcolour}{rgb}{0.97, 0.97, 0.97}
\definecolor{codeblue}{rgb}{0,0.3,0.6}

\lstdefinestyle{mystyle}{
    backgroundcolor=\color{white},   
    commentstyle=\color{codegray},
    keywordstyle=\color{codeblue},
    numberstyle=\tiny\color{codegray},
    stringstyle=\color{codeblue},
    basicstyle=\fontsize{9pt}{9.5pt}\fontfamily{lmtt}\selectfont,
    breakatwhitespace=false,         
    breaklines=true,                 
    captionpos=b,                    
    keepspaces=true,                 
    numbers=left,                    
    numbersep=12pt,                  
    showspaces=false,                
    showstringspaces=false,
    showtabs=false,                  
    tabsize=2,
    framexleftmargin=20pt,
    framextopmargin=10pt,
    framesep=30pt,
}

\usepackage[utf8]{inputenc} %
\usepackage[T1]{fontenc}    %
\usepackage{amsfonts}       %
\usepackage{nicefrac}       %
\usepackage{microtype}      %

\usepackage{epsfig}
\usepackage{graphicx}
\usepackage{xcolor}
\usepackage{amsmath}
\usepackage{amssymb}
\usepackage{subcaption}
\usepackage{textcomp}
\usepackage{gensymb}
\usepackage{mathtools}
\usepackage{amssymb}
\usepackage{wrapfig}
\usepackage{lipsum}
\usepackage{diagbox}

\usepackage{booktabs}
\usepackage{float}
\usepackage{hhline}
\usepackage{tabularx}
\newcolumntype{Y}{>{\centering\arraybackslash}X}
\newcolumntype{L}{>{\raggedright\arraybackslash}X}
\newcolumntype{s}{>{\hsize=.8\hsize}Y}
\newcolumntype{t}{>{\hsize=.6\hsize}Y}
\newcolumntype{?}{!{\vrule width 1pt}}
\usepackage{multirow}
\restylefloat{table}
\usepackage{makecell}

\usepackage{arydshln}
\usepackage{dsfont}

\usepackage{letltxmacro}
\makeatletter
\let\oldr@@t\r@@t
\def\r@@t#1#2{%
	\setbox0=\hbox{$\oldr@@t#1{#2\,}$}\dimen0=\ht0
	\advance\dimen0-0.2\ht0
	\setbox2=\hbox{\vrule height\ht0 depth -\dimen0}%
	{\box0\lower0.4pt\box2}}
\LetLtxMacro{\oldsqrt}{\sqrt}
\renewcommand*{\sqrt}[2][\ ]{\oldsqrt[#1]{#2}}
\makeatother

\makeatletter
\newcommand{\thickhline}{%
	\noalign {\ifnum 0=`}\fi \hrule height 1pt
	\futurelet \reserved@a \@xhline
}
\newcolumntype{"}{@{\hskip\tabcolsep\vrule width 1pt\hskip\tabcolsep}}
\makeatother

\makeatletter
\newcommand*{\defeq}{\mathrel{\rlap{%
			\raisebox{0.3ex}{$\m@th\cdot$}}%
		\raisebox{-0.3ex}{$\m@th\cdot$}}%
	=}
\makeatother

\usepackage{hyperref}
\hypersetup{pagebackref=true,breaklinks=true,letterpaper=true,colorlinks,bookmarks=false,%
	pdfinfo={
    		Title={The Many Faces of Robustness: A Critical Analysis of Out-of-Distribution Generalization},
		Author={Dan Hendrycks and Steven Basart and Norman Mu and Saurav Kadavath and Frank Wang and Evan Dorundo and Rahul Desai and Tyler Zhu and Samyak Parajuli and Mike Guo and Dawn Song and Jacob Steinhardt and Justin Gilmer},
		Subject={Deep Learning, Robustness, Domain Adaptation, Distribution Shift, OOD},
		Keywords={robustness, out of distribution, distribution shift, adversarial examples, domain generalization, shape, texture, out-of-distribution, OOD, ml safety}
	}
}

\iccvfinalcopy %

\def\iccvPaperID{10942} %

\ificcvfinal\fi

\usepackage{multicol}
\usepackage{appendix}
\usepackage{cleveref}
\crefname{appsec}{Appendix}{Appendices}
\usepackage{enumitem}

\newcommand{\reviewer}[3]{
	\expandafter\newcommand\csname #1\endcsname[1]{
		\textcolor{#3}{[#2: ##1]}
	}
}
\reviewer{norman}{Norman}{green}
\definecolor{neonpurple}{rgb}{0.3,0,1}
\reviewer{dan}{Dan}{neonpurple}
\reviewer{justin}{Justin}{blue}

\title{The Many Faces of Robustness: A Critical Analysis of\\Out-of-Distribution Generalization}

\author{%
    Dan Hendrycks$^1$ \and Steven Basart$^2$\thanks{Equal contribution. \textnormal{$^1$UC Berkeley}, \textnormal{$^2$UChicago}, \textnormal{$^3$Google}. Code is available at \url{https://github.com/hendrycks/imagenet-r}.} \and Norman Mu$^{1}$\footnotemark[1] \and Saurav Kadavath$^1$ \and Frank Wang$^3$ \and Evan Dorundo$^3$ \and Rahul Desai$^1$ \and Tyler Zhu$^1$ \and Samyak Parajuli$^1$ \and Mike Guo$^1$
    \and Dawn Song$^1$ \and Jacob Steinhardt$^1$ \and Justin Gilmer$^3$
}

\begin{document}

\maketitle

\begin{abstract}
We introduce four new real-world distribution shift datasets consisting of changes in image style, image blurriness, geographic location, camera operation, and more. With our new datasets, we take stock of previously proposed methods for improving out-of-distribution robustness and put them to the test. We find that using larger models and artificial data augmentations can improve robustness on real-world distribution shifts, contrary to claims in prior work. We find improvements in artificial robustness benchmarks can transfer to real-world distribution shifts, contrary to claims in prior work.
Motivated by our observation that data augmentations can help with real-world distribution shifts, we also introduce a new data augmentation method which advances the state-of-the-art and outperforms models pretrained with $1000\times$ more labeled data. Overall we find that some methods consistently help with distribution shifts in texture and local image statistics, but these methods do not help with some other distribution shifts like geographic changes. Our results show that future research must study multiple distribution shifts simultaneously, as we demonstrate that no evaluated method consistently improves robustness.\looseness=-1

\end{abstract}

\section{Introduction}
While the research community must create robust models that generalize to new scenarios, the robustness literature \cite{dodge2017study, geirhos2020shortcut} lacks consensus on evaluation benchmarks and contains many dissonant hypotheses.
Hendrycks et al., 2020 \cite{hendrycks2020pretrained} find that many recent language models are already robust to many forms of distribution shift, while others \cite{yin2019fourier,geirhos2019} find that vision models are largely fragile and argue that data augmentation offers one solution. In contrast, other researchers \cite{taori2020when} provide results suggesting that using pretraining and improving in-distribution test set accuracy improves natural robustness, whereas other methods do not. %

Prior works have also offered various interpretations of empirical results, such as the {\it Texture Bias} hypothesis that convolutional networks are biased towards texture, harming robustness \cite{geirhos2019}.
Additionally, some authors posit a fundamental distinction between robustness on \emph{synthetic} benchmarks vs. \emph{real-world} distribution shifts, casting doubt on the generality of conclusions drawn from experiments conducted on synthetic benchmarks \cite{taori2020when}.

\begin{figure*}[t]
\vspace{-15pt}
\begin{center}
\includegraphics[width=\textwidth]{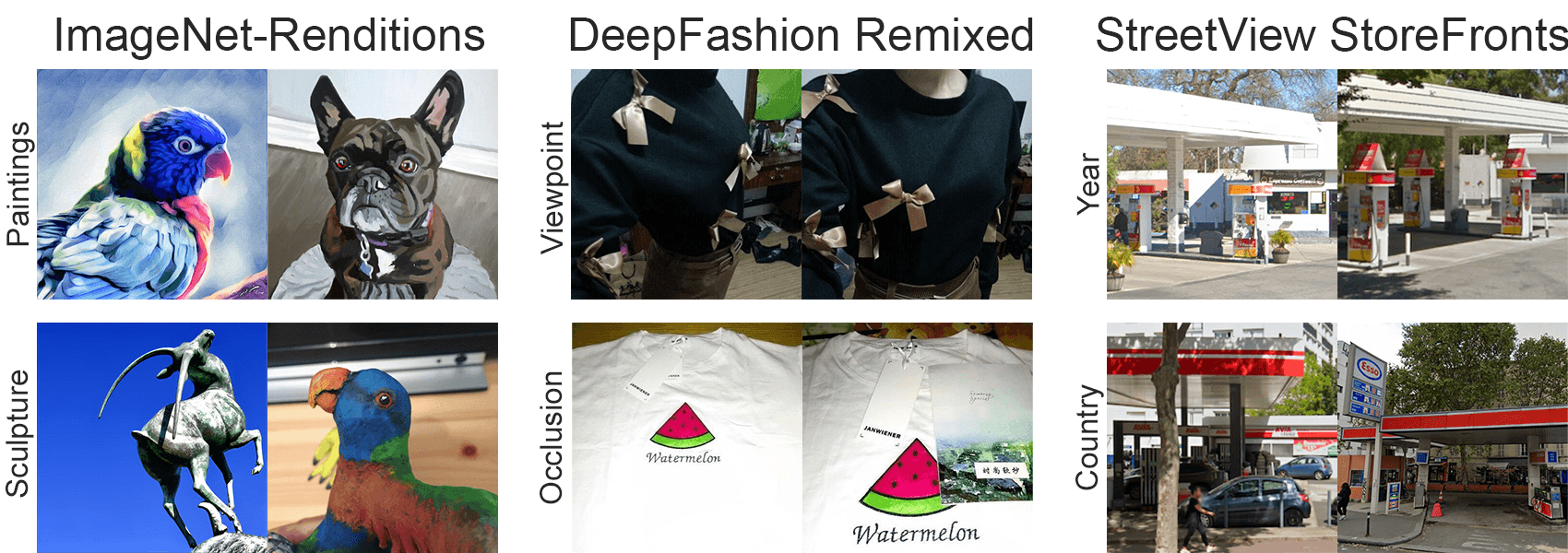}
\end{center}
\caption{
Images from three of our four new datasets ImageNet-Renditions (ImageNet-R), DeepFashion Remixed (DFR), and StreetView StoreFronts (SVSF).
The SVSF images are recreated from the public Google StreetView. %
Our datasets test robustness to various naturally occurring distribution shifts including rendition style, camera viewpoint, and geography.
}\label{fig:splash}
\vspace{-10pt}
\end{figure*}

It has been difficult to arbitrate these hypotheses because existing robustness datasets vary multiple factors (e.g., time, camera, location, etc.) simultaneously in unspecified ways \cite{Recht2019DoIC, Hendrycks2019NaturalAE}.
Existing datasets also lack diversity such that it is hard to extrapolate which methods will improve robustness more broadly. To address these issues and test the methods outlined above, we introduce four new robustness datasets and a new data augmentation method.

First we introduce ImageNet-Renditions (ImageNet-R), a 30,000 image test set containing various renditions (e.g., paintings, embroidery, etc.) of ImageNet object classes. These renditions are naturally occurring, with textures and local image statistics unlike those of ImageNet images, allowing us to compare against gains on synthetic robustness benchmarks. %

Next, we investigate the effect of changes in the image capture process with StreetView StoreFronts (SVSF) and DeepFashion Remixed (DFR). SVSF contains business storefront images collected from Google StreetView, along with metadata allowing us to vary location, year, and even the camera type. DFR leverages the metadata from DeepFashion2 \cite{ge2019deepfashion2} to systematically shift object occlusion, orientation, zoom, and scale at test time.
Both SVSF and DFR provide distribution shift controls and do not alter texture, which remove possible confounding variables affecting prior benchmarks.

Additionally, we collect Real Blurry Images, which consists of $1,\!000$ blurry natural images from a 100-class subset of the ImageNet classes. This benchmark serves as a real-world analog for the synthetic blur corruptions of the ImageNet-C benchmark \cite{hendrycks2019robustness}. With it we find that synthetic corruptions correlate with corruptions that appear in the wild, contradicting speculations from previous work \cite{taori2020when}.

Finally, we contribute DeepAugment to increase robustness to some new types of distribution shift. This augmentation technique uses image-to-image neural networks for data augmentation. DeepAugment improves robustness on our newly introduced ImageNet-R benchmark and can also be combined with other augmentation methods to outperform a model pretrained on $1000\times$ more labeled data.

We use these new datasets to test four overarching classes of methods for improving robustness:
\begin{itemize}[noitemsep,topsep=0pt,parsep=0pt,partopsep=0pt,leftmargin=12pt]
\item {\it Larger Models}: increasing model size improves robustness to distribution shift \cite{hendrycks2019robustness, Xie2020Intriguing}.
\item {\it Self-Attention}: adding self-attention layers to models improves robustness \cite{Hendrycks2019NaturalAE}.\looseness=-1
\item {\it Diverse Data Augmentation}: robustness can increase through data augmentation \cite{yin2019fourier}.
\item {\it Pretraining}: pretraining on larger and more diverse datasets improves robustness \cite{Orhan2019RobustnessPO,hendrycks2019pretrain}.
\end{itemize}

After examining our results on these four new datasets as well as prior benchmarks, we can rule out several previous hypotheses while strengthening support for others. %
As one example, we find that synthetic data augmentation robustness interventions improve accuracy on ImageNet-R and real-world image blur distribution shifts, which lends credence to the use of synthetic robustness benchmarks and also reinforces the {\it Texture Bias} hypothesis. In the conclusion, we summarize the various strands of evidence for and against each hypothesis.
Across our many experiments, we do not find a general method that consistently improves robustness, and some hypotheses require additional qualifications. While robustness is often spoken of and measured as a single scalar property like accuracy, our investigations show that robustness is not so simple. Our results show that future robustness research requires more thorough evaluation using more robustness datasets.

\section{Related Work}

\paragraph{Robustness Benchmarks.}
Recent works \cite{hendrycks2019robustness, Recht2019DoIC, hendrycks2020pretrained} have begun to characterize model performance on out-of-distribution (OOD) data with various new test sets, with dissonant findings. For instance, prior work \cite{hendrycks2020pretrained} demonstrates that modern language processing models are moderately robust to numerous naturally occurring distribution shifts, and that IID accuracy is not straightforwardly predictive of OOD accuracy for natural language tasks. For image recognition, other work \cite{hendrycks2019robustness} analyzes image models and shows that they are sensitive to various simulated image corruptions (e.g., noise, blur, weather, JPEG compression, etc.) from their ImageNet-C benchmark.

Recht et al., 2019 \cite{Recht2019DoIC} reproduce the ImageNet \cite{ILSVRC15} validation set for use as a benchmark of naturally occurring distribution shift in computer vision. Their evaluations show a 11-14\% drop in accuracy from ImageNet to the new validation set, named ImageNetV2, across a wide range of architectures. \cite{taori2020when} use ImageNetV2 to measure natural robustness and conclude that methods such as data augmentation do not significantly improve robustness. 
Recently, \cite{engstrom2020identifying} identify statistical biases in ImageNetV2's construction, and they estimate that re-weighting ImageNetV2 to correct for these biases results in a less substantial 3.6\% drop.

\begin{figure*}[t]
\vspace{-15pt}
\begin{center}
\includegraphics[width=\textwidth]{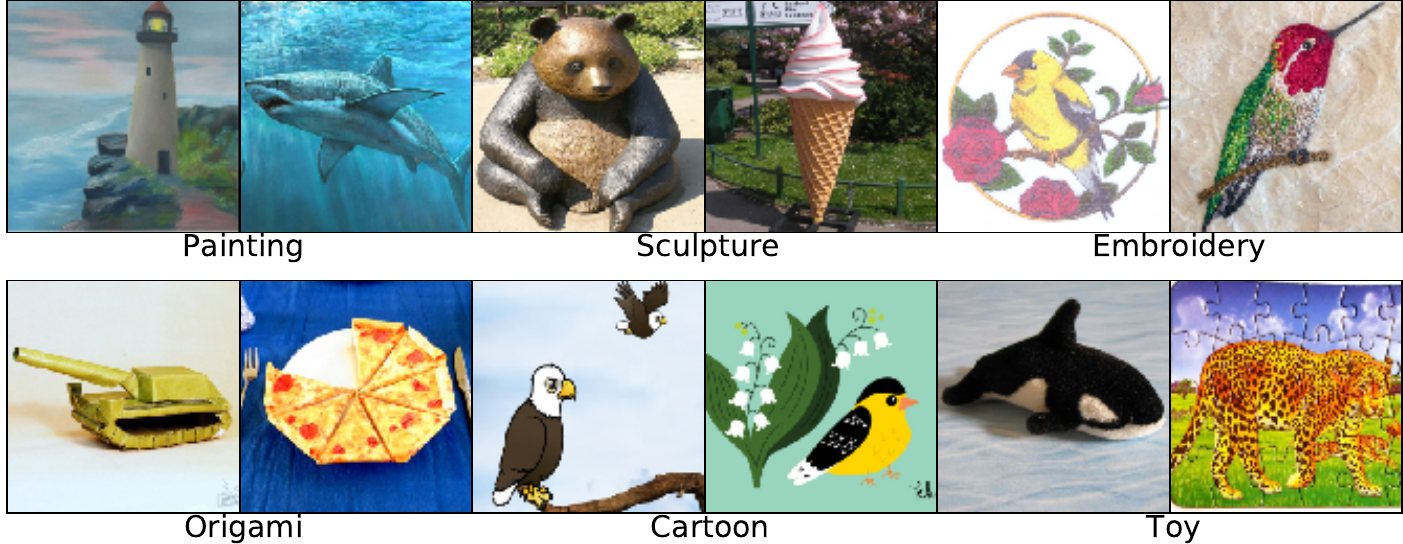}
\end{center}
\caption{
ImageNet-Renditions (ImageNet-R) contains 30,000 images of ImageNet objects with different textures and styles. This figure shows only a portion of ImageNet-R's numerous rendition styles.
The rendition styles (e.g., ``Toy'') are for clarity and are \emph{not} ImageNet-R's classes; ImageNet-R's classes are a subset of 200 ImageNet classes. %
}\label{fig:renditions}
\vspace{-10pt}
\end{figure*}

\paragraph{Data Augmentation.}

Recent works \cite{geirhos2019, yin2019fourier, hendrycks2019AugMix} demonstrate that data augmentation can improve robustness on ImageNet-C.
The space of augmentations that help robustness includes various types of noise \cite{madry2017towards, rusak2020increasing, Lopes2019ImprovingRW}, highly unnatural image transformations \cite{geirhos2019, Yun2019CutMixRS, zhang2017mixup}, or compositions of simple image transformations such as Python Imaging Library operations \cite{Cubuk2018AutoAugmentLA, hendrycks2019AugMix}.
Some of these augmentations can improve accuracy on in-distribution examples as well as on out-of-distribution (OOD) examples.

\section{New Datasets}
In order to evaluate the four robustness methods, we introduce four new benchmarks that capture new types of naturally occurring distribution shifts. %
ImageNet-Renditions (ImageNet-R) and Real Blurry Images are both newly collected test sets intended for ImageNet classifiers, whereas StreetView StoreFronts (SVSF) and DeepFashion Remixed (DFR) each contain their own training sets and multiple test sets.
SVSF and DFR split data into a training and test sets based on various image attributes stored in the metadata. For example, we can select a test set with images produced by a camera different from the training set camera. We now describe the structure and collection of each dataset.

\subsection{ImageNet-Renditions (ImageNet-R)}

While current classifiers can learn some aspects of an object's shape \cite{mordvintsev2015inceptionism}, they nonetheless rely heavily on natural textural cues \cite{geirhos2019}. In contrast, human vision can process abstract visual renditions.
For example, humans can recognize visual scenes from line drawings as quickly and accurately as they can from photographs \cite{biederman1988surface}.
Even some primates species have demonstrated the ability to recognize shape through line drawings \cite{Itakura1994,Tanaka2006}.

To measure generalization to various abstract visual renditions, we create the ImageNet-Rendition (ImageNet-R) dataset. ImageNet-R contains various artistic renditions of object classes from the original ImageNet dataset. Note the original ImageNet dataset discouraged such images since annotators were instructed to collect ``photos only, no painting, no drawings, etc.'' \cite{deng2012large}. We do the opposite.

\paragraph{Data Collection.}
ImageNet-R contains 30,000 image renditions for 200 ImageNet classes. We choose a subset of the ImageNet-1K classes, following \cite{Hendrycks2019NaturalAE}, for several reasons. A handful ImageNet classes already have many renditions, such as ``triceratops.''  We also choose a subset so that model misclassifications are egregious and to reduce label noise. The 200 class subset was also chosen based on rendition prevalence, as ``strawberry'' renditions were easier to obtain than ``radiator'' renditions. Were we to use all 1,000 ImageNet classes, annotators would be pressed to distinguish between Norwich terrier renditions as Norfolk terrier renditions, which is difficult. We collect images primarily from Flickr and use queries such as ``art,'' ``cartoon,'' ``graffiti,'' ``embroidery,'' ``graphics,'' ``origami,'' ``painting,'' ``pattern,'' ``plastic object,'' ``plush object,'' ``sculpture,'' ``line drawing,'' ``tattoo,'' ``toy,'' ``video game,'' and so on. Images are filtered by Amazon MTurk annotators using a modified collection interface from ImageNetV2 \cite{Recht2019DoIC}. For instance, after scraping Flickr images with the query ``lighthouse cartoon,'' we have MTurk annotators select true positive lighthouse renditions.
Finally, as a second round of quality control, graduate students manually filter the resulting images and ensure that individual images have correct labels and do not contain multiple labels. Examples are depicted in \Cref{fig:renditions}. %
ImageNet-R also includes the line drawings from \cite{wang2019learning}, excluding horizontally mirrored duplicate images, pitch black images, and images from the incorrectly collected ``pirate ship'' class.

\begin{figure*}[ht]
\vspace{-15pt}
\begin{center}
\includegraphics[width=\textwidth]{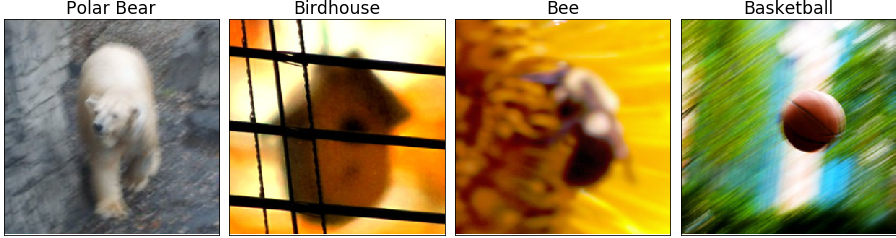}
\end{center}
\caption{
Examples of images from Real Blurry Images. This dataset allows us to test whether model performance on ImageNet-C's synthetic blur corruptions track performance on real-world blur corruptions.
}\label{fig:blur}
\vspace{-10pt}
\end{figure*}

\subsection{StreetView StoreFronts (SVSF)}

Computer vision applications often rely on data from complex pipelines that span different hardware, times, and geographies. Ambient variations in this pipeline may result in unexpected performance degradation, such as degradations experienced by health care providers in Thailand deploying laboratory-tuned diabetic retinopathy classifiers in the field \cite{beede2020human}. In order to study the effects of shifts in the image capture process we collect the StreetView StoreFronts (SVSF) dataset, a new image classification dataset sampled from Google StreetView imagery \cite{anguelov2010google} focusing on three distribution shift sources: country, year, and camera.

\paragraph{Data Collection.}

SVSF consists of cropped images of business store fronts extracted from StreetView images by an object detection model.
Each store front image is assigned the class label of the associated Google Maps business listing through a combination of machine learning models and human annotators.
We combine several visually similar business types (e.g. drugstores and pharmacies) for a total of 20 classes, listed in the Supplementary Materials. %

Splitting the data along the three metadata attributes of country, year, and camera, we create one training set and five test sets.
We sample a training set and an in-distribution test set (200K and 10K images, respectively) from images taken in US/Mexico/Canada during 2019 using a ``new'' camera system.
We then sample four OOD test sets (10K images each) which alter one attribute at a time while keeping the other two attributes consistent with the training distribution.
Our test sets are year: 2017, 2018; country: France; and camera: ``old.''

\subsection{DeepFashion Remixed}

Changes in day-to-day camera operation can cause shifts in attributes such as object size, object occlusion, camera viewpoint, and camera zoom. To measure this, we repurpose DeepFashion2 \cite{ge2019deepfashion2} to create the DeepFashion Remixed (DFR) dataset.
We designate a training set with 48K images and create eight out-of-distribution test sets to measure performance under shifts in object size, object occlusion, camera viewpoint, and camera zoom-in. DeepFashion Remixed is a multi-label classification task since images may contain more than one clothing item per image.

\paragraph{Data Collection.}

Similar to SVSF, we fix one value for each of the four metadata attributes in the training distribution.
Specifically, the DFR training set contains images with medium scale, medium occlusion, side/back viewpoint, and no zoom-in.
After sampling an IID test set, we construct eight OOD test distributions by altering one attribute at a time, obtaining test sets with minimal and heavy occlusion; small and large scale; frontal and not-worn viewpoints; and medium and large zoom-in.
See the Supplementary Materials for details on test set sizes. %

\subsection{Real Blurry Images}
We collect a small dataset of 1,000 real-world blurry images to capture real-world corruptions and validate synthetic image corruption benchmarks such as ImageNet-C. We collect the ``Real Blurry Images'' dataset from Flickr and query ImageNet object class names concatenated with the word ``blurry.'' Examples are in \Cref{fig:blur}. Each image belongs to one of 100 ImageNet classes.

\begin{figure*}[t]
\vspace{-30pt}
\begin{center}
\includegraphics[width=\textwidth]{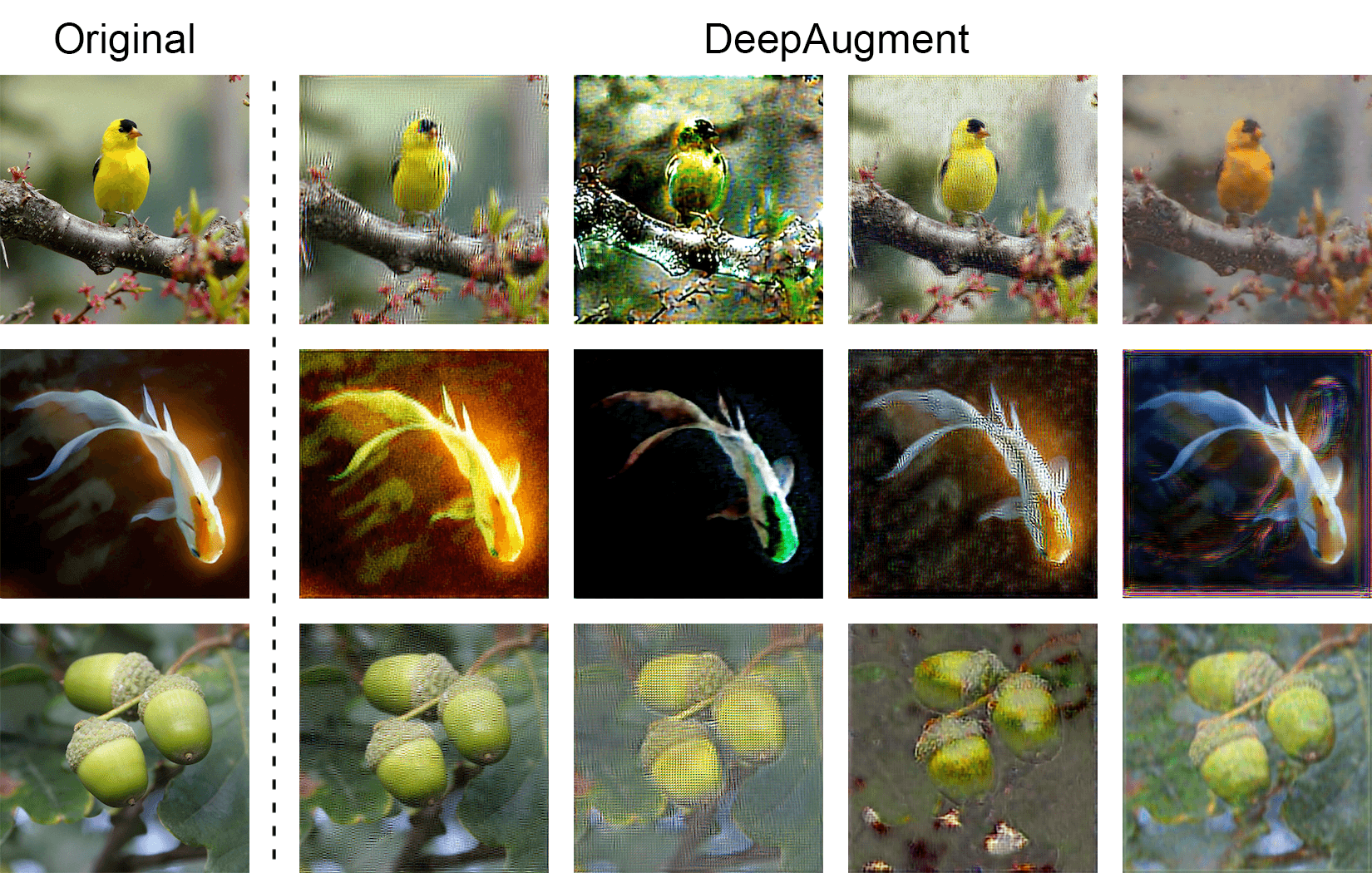}
\end{center}
\caption{
DeepAugment examples preserve semantics, are data-dependent, and are far more visually diverse than, say, rotations.\looseness=-1 %
}\label{fig:deepaugment}
\vspace{-10pt}
\end{figure*}

\section{DeepAugment}

In order to further explore effects of data augmentation, we introduce a new data augmentation technique. Whereas most previous data augmentations techniques use simple augmentation primitives applied to the raw image itself, we introduce DeepAugment, which distorts images by perturbing internal representations of deep networks.

DeepAugment works by passing a clean image through an image-to-image network and introducing several perturbations during the forward pass. These perturbations are randomly sampled from a set of manually designed functions and applied to the network weights and to the feed-forward signal at random layers. For example, our set of perturbations includes zeroing, negating, convolving, transposing, applying activation functions, and more. This setup generates semantically consistent images with unique and diverse distortions as shown in \Cref{fig:deepaugment}. Although our set of perturbations is designed with random operations, we show that DeepAugment still outperforms other methods on benchmarks such as ImageNet-C and ImageNet-R. We provide the pseudocode in the Supplementary Materials. %

For our experiments, we specifically use the CAE \cite{theis2017lossy} and EDSR \cite{lim2017enhanced} architectures as the basis for DeepAugment. CAE is an autoencoder architecture, and EDSR is a superresolution architecture. These two architectures show the DeepAugment approach works with different architectures. Each clean image in the original dataset and passed through the network and is thereby stochastically distored, resulting in two distorted versions of the clean dataset (one for CAE and one for EDSR). We then train on the augmented and clean data simultaneously and call this approach DeepAugment. The EDSR and CAE architectures are arbitrary. We show that the DeepAugment approach also works for untrained, randomly sampled architectures in the Supplementary Materials.

\section{Experiments}
\subsection{Setup}

In this section we briefly describe the evaluated models, pretraining techniques, self-attention mechanisms, data augmentation methods, and note various implementation details.

\begin{table*}[t]
\vspace{-15pt}
\begin{center}
\begin{tabular}{lcc>{\color{gray}}c}
\toprule
                        & ImageNet-200 (\%) & ImageNet-R (\%) & Gap      \\ \midrule
ResNet-50               &  7.9         &  63.9       &  56.0    \\
+ ImageNet-21K \emph{Pretraining} ($10\times$ labeled data)          &  7.0                  &  62.8             &  55.8     \\
+ CBAM (\emph{Self-Attention})                        &  7.0                  &  63.2             &  56.2     \\
+ $\ell_\infty$ Adversarial Training    &  25.1        &  68.6       &  43.5    \\
+ Speckle Noise              &  8.1                  &  62.1             &  54.0     \\
+ Style Transfer Augmentation          &  8.9         &  58.5       &  49.6    \\ 
+ AugMix                  &  7.1         &  58.9       &  51.8    \\
+ DeepAugment             &  7.5         &  57.8       &  50.3    \\
+ DeepAugment + AugMix      &  8.0         &  53.2 &  45.2 \\\midrule
ResNet-152 (\emph{Larger Models})                             &  6.8                  &  58.7             &  51.9     \\
\bottomrule
\end{tabular}
\end{center}
\caption{ImageNet-200 and ImageNet-R top-1 error rates. ImageNet-200 uses the same 200 classes as ImageNet-R. DeepAugment+AugMix improves over the baseline by over 10 percentage points. We take ImageNet-21K Pretraining and CBAM as representatives of pretraining and self-attention, respectively. Style Transfer, AugMix, and DeepAugment are all instances of more complex data augmentation, in contrast to simpler noise-based augmentations such as $\ell_\infty$ Adversarial Noise and Speckle Noise. While there remains much room for improvement, results indicate that progress on ImageNet-R is tractable.}
\label{tab:imagenetr}
\vspace{-10pt}
\end{table*}

\noindent\textbf{Model Architectures and Sizes.}\quad Most experiments are evaluated on a standard ResNet-50 model \cite{he2015deep}. Model size evaluations use ResNets or ResNeXts \cite{xie2016aggregated} of varying sizes.

\noindent\textbf{Pretraining.}\quad 
For pretraining we use ImageNet-21K which contains approximately 21,000 classes and approximately 14 million labeled training images, or around $10\times$ more labeled training data than ImageNet-1K. We also tune an ImageNet-21K model \cite{kolesnikov2019large}.
We also use a large pre-trained ResNeXt-101 model \cite{instagram2018}. This was pre-trained on on approximately 1 billion Instagram images with hashtag labels and fine-tuned on ImageNet-1K. This Weakly Supervised Learning (WSL) pretraining strategy uses approximately $1000\times$ more labeled data.

\noindent\textbf{Self-Attention.}\quad
When studying self-attention, we employ CBAM \cite{woo2018cbam} and SE \cite{Hu2018SqueezeandExcitationN} modules, two forms of self-attention that help models learn spatially distant dependencies.

\noindent\textbf{Data Augmentation.}\quad
We use Style Transfer, AugMix, and DeepAugment to evaluate the benefits of data augmentation, and we contrast their performance with simpler noise augmentations such as Speckle Noise and adversarial noise.
Style transfer \cite{geirhos2019} uses a style transfer network to apply artwork styles to training images. We use AugMix \cite{hendrycks2019AugMix} which randomly composes simple augmentation operations (e.g., translate, posterize, solarize). DeepAugment, introduced above, distorts the weights and feedforward passes of image-to-image models to generate image augmentations. Speckle Noise data augmentation muliplies each pixel by $(1+x)$ with $x$ sampled from a normal distribution \cite{rusak2020increasing,hendrycks2019robustness}. We also consider adversarial training as a form of adaptive data augmentation and use the model from \cite{wong2020fast} trained against $\ell_\infty$ perturbations of size $\varepsilon = 4/255$.

\subsection{Results}
We now perform experiments on ImageNet-R, StreetView StoreFronts, DeepFashion Remixed, and Real Blurry Images. We also evaluate on ImageNet-C and compare and contrast it with real distribution shifts.

\begin{table*}[h]
\vspace{-10pt}
\begin{center}
{%
\begin{tabular}{@{}l ? c | c | c c | c}%
\multicolumn{2}{c}{} & \multicolumn{1}{c}{Hardware} & \multicolumn{2}{c}{Year} & \multicolumn{1}{c}{Location} \\
\hline
Network & \multicolumn{1}{c|}{\,IID\,}
    & Old & 2017 & 2018 & France %
    \\ \hline 
ResNet-50               & 27.2 & 28.6 & 27.7 & 28.3 & 56.7 \\
+ Speckle Noise         & 28.5 & 29.5 & 29.2 & 29.5 & 57.4 \\
+ Style Transfer        & 29.9 & 31.3 & 30.2 & 31.2 & 59.3 \\
+ DeepAugment           & 30.5 & 31.2 & 30.2 & 31.3 & 59.1\\
+ AugMix                & 26.6 & 28.0 & 26.5 & 27.7 & 55.4 \\

\Xhline{2\arrayrulewidth}
\end{tabular}}
\caption{SVSF classification error rates. Networks are robust to some natural distribution shifts but are substantially more sensitive than the geographic shift. Here data augmentation hardly helps.}
\vspace{-10pt}
\label{tab:svsf}
\end{center}
\end{table*}

\paragraph{ImageNet-R.}
\Cref{tab:imagenetr} shows performance on ImageNet-R as well as on ImageNet-200 (the original ImageNet data restricted to ImageNet-R's 200 classes). This has several implications regarding the four method-specific hypotheses. Pretraining with ImageNet-21K (approximately $10\times$ labeled data) hardly helps. The Supplementary Materials %
shows WSL pretraining can help, but Instagram has renditions, while ImageNet excludes them; hence we conclude comparable pretraining was ineffective. Notice self-attention increases the IID/OOD gap. Compared to simpler data augmentation techniques such as Speckle Noise, the data augmentation techniques of Style Transfer, AugMix, and DeepAugment improve generalization. Note AugMix and DeepAugment improve in-distribution performance whereas Style transfer hurts it. Also, our new DeepAugment technique is the best standalone method with an error rate of 57.8\%.
Last, larger models reduce the IID/OOD gap.

As for prior hypothesis in the literature regarding model robustness, we find that
biasing networks away from natural textures through diverse data augmentation improved performance.
The IID/OOD generalization gap varies greatly by method, demonstrating that it is possible to significantly outperform the trendline of models optimized solely for the IID setting. Finally, as  ImageNet-R contains real-world examples, and since data augmentation helps on ImageNet-R, we now have clear evidence against the hypothesis that robustness interventions cannot help with natural distribution shifts \cite{taori2020when}.

\noindent\textbf{StreetView StoreFronts.}\quad
In \Cref{tab:svsf}, we evaluate data augmentation methods on SVSF and find that all of the tested methods have mostly similar performance and that no method helps much on country shift, where error rates roughly double across the board. Here evaluation is limited to augmentations due to a 30 day retention window for each instantiation of the dataset. Images captured in France contain noticeably different architectural styles and storefront designs than those captured in US/Mexico/Canada; meanwhile, we are unable to find conspicuous and consistent indicators of the camera and year.
This may explain the relative insensitivity of evaluated methods to the camera and year shifts.
Overall data augmentation here shows limited benefit, suggesting either that data augmentation primarily helps combat texture bias as with ImageNet-R, or that existing augmentations are not diverse enough to capture high-level semantic shifts such as building architecture.

\begin{table*}[h]
\footnotesize
\begin{center}
{\setlength\tabcolsep{2pt}%
\begin{tabular}{@{}l | c | c | c c | c c | c c | c c}%
\multicolumn{3}{c}{} & \multicolumn{2}{c}{Size} & \multicolumn{2}{c}{Occlusion} & \multicolumn{2}{c}{Viewpoint} & \multicolumn{2}{c}{Zoom} \\
\hline
Network & \multicolumn{1}{c|}{\,IID\,} & \multicolumn{1}{c|}{\,OOD\,}
& Small & Large 
& Slight/None & Heavy
& No Wear & Side/Back
& Medium & Large
    \\ \hline 
ResNet-50	&	77.6    & 55.1	&	39.4	&	73.0	&	51.5	&	41.2	&	50.5	&	63.2	&	48.7	&	73.3	\\
+ ImageNet-21K \emph{Pretraining} &  80.8   & 58.3  & 40.0  & 73.6  &  55.2 & 43.0  &  63.0 & 67.3  &  50.5 &  73.9 \\
+ SE (\emph{Self-Attention})  & 77.4  & 55.3 &  38.9 & 72.7 & 52.1  & 40.9  & 52.9  & 64.2  & 47.8  & 72.8 \\
+ Random Erasure    & 78.9  &  56.4 &  39.9  &  75.0 &  52.5 &  42.6 &  53.4 &  66.0 & 48.8  &  73.4 \\
+ Speckle Noise	&	78.9	& 55.8  &	38.4	&  74.0 	&	52.6	&	40.8	&	55.7	&	63.8	&	47.8	&	73.6	\\
+ Style Transfer	&	80.2	&   57.1   &	37.6	&	76.5	&	54.6	&	43.2	&	58.4	&	65.1	&	49.2	&	72.5	\\
+ DeepAugment	&	79.7	&   56.3   &	38.3	&	74.5	&	52.6	&	42.8	&	54.6	&	65.5	&	49.5	&	72.7	\\
+ AugMix	&	80.4	&   57.3   &	39.4	&	74.8	&	55.3	&	42.8	&	57.3	&	66.6	&	49.0	&	73.1	\\ \midrule
ResNet-152 (\emph{Larger Models})  & 80.0  &    57.1   & 40.0  & 75.6 & 52.3  & 42.0 &  57.7 &  65.6 &  48.9 & 74.4  \\
\Xhline{2\arrayrulewidth}
\end{tabular}}
\end{center}
\caption{DeepFashion Remixed results. Unlike the previous tables, higher is better since all values are mAP scores for this multi-label classification benchmark. The ``OOD'' column is the average of the row's rightmost eight OOD values. All techniques do little to close the IID/OOD generalization gap. %
}
\vspace{-10pt}
\label{tab:deepfashion}
\end{table*}

\begin{table*}[t]
\setlength{\tabcolsep}{12pt}
\centering
\begin{tabular}{lcccccc}
  Method       & ImageNet-C & Real Blurry Images & ImageNet-R  & DFR \\ 
\hline
Larger Models    &  $+$ & $+$ & $+$ & $-$ \\
Self-Attention   & $+$ & $+$ & $-$ & $-$  \\
Diverse Data Augmentation       & $+$ & $+$ & $+$ & $-$  \\
Pretraining       & $+$ & $+$ & $-$ & $-$  \\
\bottomrule
\end{tabular}
\caption{A highly simplified account of each method when tested against different datasets. Evidence for is denoted ``$+$'', and ``$-$'' denotes an absence of evidence or evidence against.}
\label{tab:hypothesissummary}
\end{table*}

\noindent \textbf{DeepFashion Remixed.}\quad
\Cref{tab:deepfashion} shows our experimental findings on DFR, in which all evaluated methods have an average OOD mAP that is close to the baseline. In fact, most OOD mAP increases track IID mAP increases. In general, DFR's size and occlusion shifts hurt performance the most. We also evaluate with Random Erasure augmentation, which deletes rectangles within the image, to simulate occlusion \cite{Zhong2017RandomED}.
Random Erasure improved occlusion performance, but Style Transfer helped even more.
Nothing substantially improved OOD performance beyond what is explained by IID performance, so here it would appear that in this setting, only IID performance matters.
Our results suggest that while some methods may improve robustness to certain forms of distribution shift, no method substantially raises performance across all shifts.

\begin{figure}
	\centering
      \includegraphics[width=\linewidth]{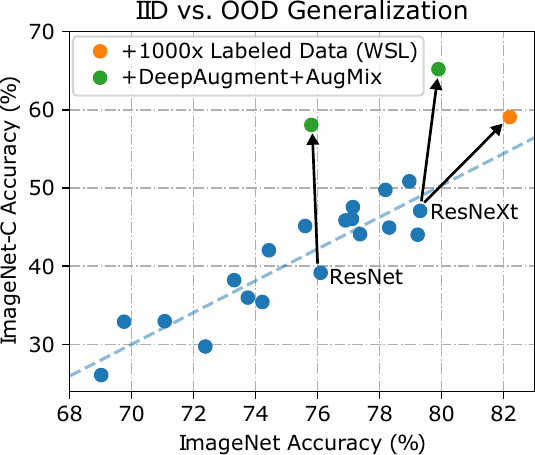}
  \caption{ImageNet accuracy and ImageNet-C accuracy. Previous architectural advances slowly translate to ImageNet-C performance improvements, but DeepAugment+AugMix on a ResNet-50 yields approximately a $19\%$ accuracy increase. This shows IID accuracy and OOD accuracy are not coupled, contra \cite{taori2020when}. %
  }
  \label{fig:imagenetc}
\end{figure}

\noindent\textbf{Real Blurry Images and ImageNet-C.}\quad
We now consider a previous robustness benchmark to evaluate the four major methods. We use the ImageNet-C dataset \cite{hendrycks2019robustness} which applies 15 common image corruptions (e.g., Gaussian noise, defocus blur, simulated fog, JPEG compression, etc.) across 5 severities to ImageNet-1K validation images. We find that DeepAugment improves robustness on ImageNet-C. \Cref{fig:imagenetc} shows that when models are trained with both AugMix and DeepAugment they set a new state-of-the-art, breaking the trendline and exceeding the corruption robustness provided by training on $1000\times$ more labeled training data. Note the augmentations from AugMix and DeepAugment are disjoint from ImageNet-C's corruptions. Full results are shown in the Supplementary Materials. %
IID accuracy alone is clearly unable to capture the full story of model robustness. Instead, larger models, self-attention, data augmentation, and pretraining all improve robustness far beyond the degree predicted by their influence on IID accuracy.

A recent work \cite{taori2020when} reminds us that ImageNet-C uses various \emph{synthetic} corruptions and suggest that they are decoupled from real-world robustness. Real-world robustness requires generalizing to naturally occurring corruptions such as snow, fog, blur, low-lighting noise, and so on, but it is an open question whether ImageNet-C's simulated corruptions meaningfully approximate real-world corruptions.

We evaluate various models on Real Blurry Images and find that \emph{all} the robustness interventions that help with ImageNet-C also help with real-world blurry images. Hence ImageNet-C can track performance on real-world corruptions. Moreover, DeepAugment+AugMix has the lowest error rate on Real Blurry Images, which again contradicts the synthetic vs natural dichotomy. The upshot is that ImageNet-C is a controlled and systematic proxy for real-world robustness.

Our results, which are expanded on in the Supplementary Materials, show that larger models, self-attention, data augmentation, and pretraining all help, just like on ImageNet-C. Here DeepAugment+AugMix attains state-of-the-art. These results suggest ImageNet-C's simulated corruptions track real-world corruptions. In hindsight, this is expected since various computer vision problems have used synthetic corruptions as proxies for real-world corruptions, for decades. In short, ImageNet-C is a diverse and systematic benchmark that is correlated with improvements on real-world corruptions.

\section{Conclusion}

In this paper we introduced four real-world datasets for evaluating the robustness of computer vision models: ImageNet-Renditions, DeepFashion Remixed, StreetView StoreFronts, and Real Blurry Images.
With our new datasets, we re-evaluate previous robustness interventions and determine whether various robustness hypotheses are correct or incorrect in view of our new findings.

Our main results for different robustness interventions are as follows. Larger models improved robustness on Real Blurry Images, ImageNet-C, and ImageNet-R, but not with DFR.
While self-attention noticeably helped Real Blurry Images and ImageNet-C, it did not help with ImageNet-R and DFR.
Diverse data augmentation was ineffective for SVSF and DFR, but it greatly improved accuracy on Real Blurry Images, ImageNet-C, and ImageNet-R.
Pretraining greatly helped with Real Blurry Images and ImageNet-C but hardly helped with DFR and ImageNet-R. It was not obvious \emph{a priori} that synthetic data augmentation could improve accuracy on a real-world distribution shift such as ImageNet-R, nor had pretraining ever failed to improve performance in earlier research \cite{taori2020when}. Table~\ref{tab:hypothesissummary} shows that many methods improve robustness across multiple distribution shifts. While no single method consistently helped across all distribution shifts, some helped more than others.

Our analysis also has implications for the three robustness hypotheses. In support of the \emph{Texture Bias} hypothesis, ImageNet-R shows that standard networks do not generalize well to renditions (which have different textures), but that diverse data augmentation (which often distorts textures) can recover accuracy. More generally, larger models and diverse data augmentation consistently helped on ImageNet-R, ImageNet-C, and Real Blurry Images, suggesting that these two interventions reduce texture bias. However, these methods helped little for geographic shifts, showing that there is more to robustness than texture bias alone. Regarding more general trends across the last several years of progress in deep learning, while IID accuracy is a strong predictor of OOD accuracy, it is not decisive, contrary to some prior works \cite{taori2020when}. %
Again contrary to a hypothesis from prior work \cite{taori2020when}, our findings show that the gains from data augmentation on ImageNet-C generalize to both ImageNet-R and Real Blurry Images serve as a resounding validation of using synthetic benchmarks to measure model robustness.

The existing literature presents several conflicting accounts of robustness. What led to this conflict?
We suspect that this is due in large part to inconsistent notions of how to best evaluate robustness, and in particular a desire to simplify the problem by establishing the primacy of a single benchmark over others.
In response, we collected several additional datasets which each capture new dimensions of distribution shift and degradations in model performance not well studied before.
These new datasets demonstrate the importance of conducting multi-faceted evaluations of robustness as well as the general complexity of the landscape of robustness research, where it seems that so far nothing consistently helps in all settings. 
Hence the research community may consider prioritizing the study of new robustness methods, and we encourage the research community to evaluate future methods on multiple distribution shifts.
For example, ImageNet models should at least be tested against ImageNet-C and ImageNet-R.
By heightening experimental standards for robustness research, we facilitate future work towards developing systems that can robustly generalize in safety-critical settings.

\newpage

{\small
\bibliographystyle{ieee_fullname}
\bibliography{main}

\begin{thebibliography}{10}\itemsep=-1pt

\bibitem{anguelov2010google}
Dragomir Anguelov, Carole Dulong, Daniel Filip, Christian Frueh, St{\'e}phane
  Lafon, Richard Lyon, Abhijit Ogale, Luc Vincent, and Josh Weaver.
\newblock Google street view: Capturing the world at street level.
\newblock {\em Computer}, 43(6):32--38, 2010.

\bibitem{beede2020human}
Emma Beede, Elizabeth Baylor, Fred Hersch, Anna Iurchenko, Lauren Wilcox,
  Paisan Ruamviboonsuk, and Laura~M Vardoulakis.
\newblock A human-centered evaluation of a deep learning system deployed in
  clinics for the detection of diabetic retinopathy.
\newblock In {\em Proceedings of the 2020 CHI Conference on Human Factors in
  Computing Systems}, pages 1--12, 2020.

\bibitem{biederman1988surface}
Irving Biederman and Ginny Ju.
\newblock Surface versus edge-based determinants of visual recognition.
\newblock {\em Cognitive psychology}, 20(1):38--64, 1988.

\bibitem{Cubuk2018AutoAugmentLA}
Ekin~Dogus Cubuk, Barret Zoph, Dandelion Man{\'e}, Vijay Vasudevan, and Quoc~V.
  Le.
\newblock {AutoAugment}: Learning augmentation policies from data.
\newblock {\em CVPR}, 2018.

\bibitem{deng2012large}
Jia Deng.
\newblock Large scale visual recognition.
\newblock Technical report, Princeton, 2012.

\bibitem{ImageNet}
Jia Deng, Wei Dong, Richard Socher, Li jia Li, Kai Li, and Li Fei-Fei.
\newblock {I}mage{N}et: A large-scale hierarchical image database.
\newblock {\em CVPR}, 2009.

\bibitem{dodge2017study}
Samuel Dodge and Lina Karam.
\newblock A study and comparison of human and deep learning recognition
  performance under visual distortions.
\newblock In {\em 2017 26th international conference on computer communication
  and networks (ICCCN)}, pages 1--7. IEEE, 2017.

\bibitem{engstrom2020identifying}
Logan Engstrom, Andrew Ilyas, Shibani Santurkar, Dimitris Tsipras, Jacob
  Steinhardt, and Aleksander Madry.
\newblock Identifying statistical bias in dataset replication.
\newblock {\em ICML}, 2020.

\bibitem{Gao2019Res2NetAN}
Shanghua Gao, Ming-Ming Cheng, Kai Zhao, Xinyu Zhang, Ming-Hsuan Yang, and
  Philip H.~S. Torr.
\newblock Res2net: A new multi-scale backbone architecture.
\newblock {\em IEEE transactions on pattern analysis and machine intelligence},
  2019.

\bibitem{res2net_gao2019}
Shanghua Gao, Ming-Ming Cheng, Kai Zhao, Xin-Yu Zhang, Ming-Hsuan Yang, and
  Philip~H.S. Torr.
\newblock Res2net: A new multi-scale backbone architecture.
\newblock {\em IEEE Transactions on Pattern Analysis and Machine Intelligence},
  2019.

\bibitem{ge2019deepfashion2}
Yuying Ge, Ruimao Zhang, Xiaogang Wang, Xiaoou Tang, and Ping Luo.
\newblock Deepfashion2: A versatile benchmark for detection, pose estimation,
  segmentation and re-identification of clothing images.
\newblock In {\em Proceedings of the IEEE Conference on Computer Vision and
  Pattern Recognition}, pages 5337--5345, 2019.

\bibitem{geirhos2020shortcut}
Robert Geirhos, J{\"o}rn-Henrik Jacobsen, Claudio Michaelis, Richard Zemel,
  Wieland Brendel, Matthias Bethge, and Felix~A Wichmann.
\newblock Shortcut learning in deep neural networks.
\newblock {\em arXiv preprint arXiv:2004.07780}, 2020.

\bibitem{geirhos2019}
Robert Geirhos, Patricia Rubisch, Claudio Michaelis, Matthias Bethge, Felix~A
  Wichmann, and Wieland Brendel.
\newblock Imagenet-trained {CNNs} are biased towards texture; increasing shape
  bias improves accuracy and robustness.
\newblock {\em ICLR}, 2019.

\bibitem{he2015deep}
Kaiming He, Xiangyu Zhang, Shaoqing Ren, and Jian Sun.
\newblock Deep residual learning for image recognition. corr abs/1512.03385
  (2015), 2015.

\bibitem{hendrycks2019robustness}
Dan Hendrycks and Thomas Dietterich.
\newblock Benchmarking neural network robustness to common corruptions and
  perturbations.
\newblock {\em ICLR}, 2019.

\bibitem{hendrycks2019pretrain}
Dan Hendrycks, Kimin Lee, and Mantas Mazeika.
\newblock Using pre-training can improve model robustness and uncertainty.
\newblock In {\em ICML}, 2019.

\bibitem{hendrycks2020pretrained}
Dan Hendrycks, Xiaoyuan Liu, Eric Wallace, Adam Dziedzic, Rishabh Krishnan, and
  Dawn Song.
\newblock Pretrained transformers improve out-of-distribution robustness.
\newblock {\em ACL}, 2020.

\bibitem{hendrycks2019AugMix}
Dan Hendrycks, Norman Mu, Ekin~D Cubuk, Barret Zoph, Justin Gilmer, and Balaji
  Lakshminarayanan.
\newblock Augmix: A simple data processing method to improve robustness and
  uncertainty.
\newblock {\em ICLR}, 2020.

\bibitem{Hendrycks2019NaturalAE}
Dan Hendrycks, Kevin Zhao, Steven Basart, Jacob Steinhardt, and Dawn Song.
\newblock Natural adversarial examples.
\newblock {\em ArXiv}, abs/1907.07174, 2019.

\bibitem{Hu2018SqueezeandExcitationN}
Jie Hu, Li Shen, and Gang Sun.
\newblock Squeeze-and-excitation networks.
\newblock {\em 2018 IEEE/CVF Conference on Computer Vision and Pattern
  Recognition}, 2018.

\bibitem{Itakura1994}
Shoji Itakura.
\newblock Recognition of line-drawing representations by a chimpanzee (pan
  troglodytes).
\newblock {\em The Journal of General Psychology}, 121(3):189--197, July 1994.

\bibitem{kolesnikov2019large}
Alexander Kolesnikov, Lucas Beyer, Xiaohua Zhai, Joan Puigcerver, Jessica Yung,
  Sylvain Gelly, and Neil Houlsby.
\newblock Large scale learning of general visual representations for transfer.
\newblock {\em arXiv preprint arXiv:1912.11370}, 2019.

\bibitem{Lee2020NetworkRA}
Kimin Lee, Kibok Lee, Jinwoo Shin, and Honglak Lee.
\newblock Network randomization: A simple technique for generalization in deep
  reinforcement learning.
\newblock In {\em ICLR}, 2020.

\bibitem{lim2017enhanced}
Bee Lim, Sanghyun Son, Heewon Kim, Seungjun Nah, and Kyoung Mu~Lee.
\newblock Enhanced deep residual networks for single image super-resolution.
\newblock In {\em Proceedings of the IEEE conference on computer vision and
  pattern recognition workshops}, pages 136--144, 2017.

\bibitem{Lopes2019ImprovingRW}
Raphael~Gontijo Lopes, Dong Yin, Ben Poole, Justin Gilmer, and Ekin~Dogus
  Cubuk.
\newblock Improving robustness without sacrificing accuracy with patch
  {Gaussian} augmentation.
\newblock {\em arXiv preprint arXiv:1906.02611}, 2019.

\bibitem{madry2017towards}
Aleksander Madry, Aleksandar Makelov, Ludwig Schmidt, Dimitris Tsipras, and
  Adrian Vladu.
\newblock Towards deep learning models resistant to adversarial attacks.
\newblock {\em arXiv preprint arXiv:1706.06083}, 2017.

\bibitem{instagram2018}
Dhruv Mahajan, Ross Girshick, Vignesh Ramanathan, Kaiming He, Manohar~Paluri
  abd Yixuan~Li, Ashwin Bharambe, and Laurens van~der Maaten.
\newblock Exploring the limits of weakly supervised pretraining.
\newblock {\em ECCV}, 2018.

\bibitem{mordvintsev2015inceptionism}
Alexander Mordvintsev, Christopher Olah, and Mike Tyka.
\newblock Inceptionism: Going deeper into neural networks.
\newblock {\em arXiv}, 2015.

\bibitem{Orhan2019RobustnessPO}
A.~Emin Orhan.
\newblock Robustness properties of facebook's {ResNeXt WSL} models.
\newblock {\em ArXiv}, abs/1907.07640, 2019.

\bibitem{Recht2019DoIC}
Benjamin Recht, Rebecca Roelofs, Ludwig Schmidt, and Vaishaal Shankar.
\newblock Do {ImageNet} classifiers generalize to {ImageNet}?
\newblock {\em ArXiv}, abs/1902.10811, 2019.

\bibitem{rusak2020increasing}
Evgenia Rusak, Lukas Schott, Roland Zimmermann, Julian Bitterwolf, Oliver
  Bringmann, Matthias Bethge, and Wieland Brendel.
\newblock Increasing the robustness of dnns against image corruptions by
  playing the game of noise.
\newblock {\em arXiv preprint arXiv:2001.06057}, 2020.

\bibitem{ILSVRC15}
Olga Russakovsky, Jia Deng, Hao Su, Jonathan Krause, Sanjeev Satheesh, Sean Ma,
  Zhiheng Huang, Andrej Karpathy, Aditya Khosla, Michael Bernstein,
  Alexander~C. Berg, and Li Fei-Fei.
\newblock {ImageNet Large Scale Visual Recognition Challenge}.
\newblock {\em International Journal of Computer Vision (IJCV)},
  115(3):211--252, 2015.

\bibitem{Tanaka2006}
Masayuki Tanaka.
\newblock Recognition of pictorial representations by chimpanzees (pan
  troglodytes).
\newblock {\em Animal Cognition}, 10(2):169--179, Dec. 2006.

\bibitem{taori2020when}
Rohan Taori, Achal Dave, Vaishaal Shankar, Nicholas Carlini, Benjamin Recht,
  and Ludwig Schmidt.
\newblock When robustness doesn{\textquoteright}t promote robustness: Synthetic
  vs. natural distribution shifts on imagenet, 2020.

\bibitem{theis2017lossy}
Lucas Theis, Wenzhe Shi, Andrew Cunningham, and Ferenc Husz{\'a}r.
\newblock Lossy image compression with compressive autoencoders.
\newblock {\em arXiv preprint arXiv:1703.00395}, 2017.

\bibitem{Wang2020I}
Haotao Wang, Tianlong Chen, Zhangyang Wang, and Kede Ma.
\newblock I am going mad: Maximum discrepancy competition for comparing
  classifiers adaptively.
\newblock In {\em International Conference on Learning Representations}, 2020.

\bibitem{wang2019learning}
Haohan Wang, Songwei Ge, Eric~P. Xing, and Zachary~C. Lipton.
\newblock Learning robust global representations by penalizing local predictive
  power, 2019.

\bibitem{wong2020fast}
Eric Wong, Leslie Rice, and J~Zico Kolter.
\newblock Fast is better than free: Revisiting adversarial training.
\newblock {\em arXiv preprint arXiv:2001.03994}, 2020.

\bibitem{woo2018cbam}
Sanghyun Woo, Jongchan Park, Joon-Young Lee, and In So~Kweon.
\newblock Cbam: Convolutional block attention module.
\newblock In {\em Proceedings of the European Conference on Computer Vision
  (ECCV)}, pages 3--19, 2018.

\bibitem{Xie2020Intriguing}
Cihang Xie and Alan Yuille.
\newblock Intriguing properties of adversarial training at scale.
\newblock In {\em International Conference on Learning Representations}, 2020.

\bibitem{xie2016aggregated}
Saining Xie, Ross Girshick, Piotr Doll{\'a}r, Zhuowen Tu, and Kaiming He.
\newblock Aggregated residual transformations for deep neural networks. 2016.
\newblock {\em arXiv preprint arXiv:1611.05431}, 2016.

\bibitem{yin2019fourier}
Dong Yin, Raphael~Gontijo Lopes, Jonathon Shlens, Ekin~D Cubuk, and Justin
  Gilmer.
\newblock A {F}ourier perspective on model robustness in computer vision.
\newblock {\em arXiv preprint arXiv:1906.08988}, 2019.

\bibitem{Yun2019CutMixRS}
Sangdoo Yun, Dongyoon Han, Seong~Joon Oh, Sanghyuk Chun, Junsuk Choe, and
  Youngjoon Yoo.
\newblock Cutmix: Regularization strategy to train strong classifiers with
  localizable features.
\newblock {\em ICCV}, 2019.

\bibitem{zhang2017mixup}
Hongyi Zhang, Moustapha Cisse, Yann~N Dauphin, and David Lopez-Paz.
\newblock mixup: Beyond empirical risk minimization.
\newblock {\em arXiv preprint arXiv:1710.09412}, 2017.

\bibitem{zhang2018perceptual}
Richard Zhang, Phillip Isola, Alexei~A Efros, Eli Shechtman, and Oliver Wang.
\newblock The unreasonable effectiveness of deep features as a perceptual
  metric.
\newblock In {\em CVPR}, 2018.

\bibitem{Zhong2017RandomED}
Zhun Zhong, Liang Zheng, Guoliang Kang, Shaozi Li, and Yi Yang.
\newblock Random erasing data augmentation.
\newblock {\em arXiv preprint arXiv:1708.04896}, 2017.

\end{thebibliography}
}

\newpage
\appendix
\newpage
\begin{appendices}

\section{Additional Results}\label{app:additional}

\noindent \textbf{ImageNet-R.}\quad
Expanded ImageNet-R results are in \Cref{tab:imagenetr_full}.
WSL pretraining on Instagram images appears to yield dramatic improvements on ImageNet-R, but the authors note the prevalence of artistic renditions of object classes on the Instagram platform. While ImageNet's data collection process actively excluded renditions, we do not have reason to believe the Instagram dataset excluded renditions. On a ResNeXt-101 32$\times$8d model, WSL pretraining improves ImageNet-R performance by a massive 37.5\% from 57.5\% top-1 error to 24.2\%. Ultimately, without examining the training images we are unable to determine whether ImageNet-R represents an actual distribution shift to the Instagram WSL models. However, we also observe that with greater controls, that is with ImageNet-21K pre-training, pretraining hardly helped ImageNet-R performance, so it is not clear that more pretraining data improves ImageNet-R performance.

\begin{figure}
	\begin{center}
	\includegraphics[width=0.4\textwidth]{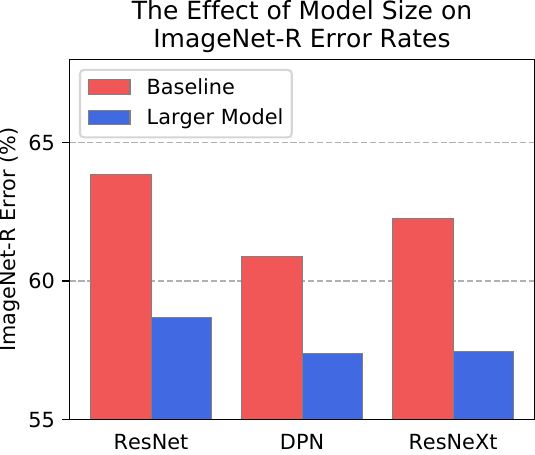}
	\end{center}
	\caption{
	Larger models improve robustness on ImageNet-R. The baseline models are ResNet-50, DPN-68, and ResNeXt-50 ($32\times4$d). The larger models are ResNet-152, DPN-98, and ResNeXt-101 ($32\times8$d). The baseline ResNeXt has a 7.1\% ImageNet error rate, while the large has a 6.2\% error rate.
	}\label{fig:imagenetr}
\end{figure}

Increasing model size appears to automatically improve ImageNet-R performance, as shown in \Cref{fig:imagenetr}. A ResNet-50 (25.5M parameters) has 63.9\% error, while a ResNet-152 (60M) has 58.7\% error. ResNeXt-50 32$\times$4d (25.0M) attains 62.3\% error and ResNeXt-101 32$\times$8d (88M) attains 57.5\% error.

\noindent \textbf{ImageNet-C.}\quad
Expanded ImageNet-C results are \Cref{tab:imagenetc_table}. We also tested whether model size improves performance on ImageNet-C for even larger models. With a different codebase, we trained ResNet-50, ResNet-152, and ResNet-500 models which achieved 80.6, 74.0, and 68.5 mCE respectively. Expanded comparisons between ImageNet-C and Real Blurry Images is in \Cref{tab:imagenetblur}.

\begin{table*}[t]
\vspace{-10pt}
\centering
\begin{tabularx}{\textwidth}{*{1}{>{\hsize=1.4\hsize}L} | *{4}{>{\hsize=0.25\hsize}Y} || *{1}{>{\hsize=0.6\hsize}Y} | *{1}{>{\hsize=0.6\hsize}Y}}%
Network & Defocus Blur & Glass Blur & Motion Blur & Zoom Blur & {ImageNet-C Blur Mean} & Real Blurry Images
    \\ \hline 
ResNet-50	& 61 & 73 & 61 & 64 & 65 & 58.7 \\
+ ImageNet-21K \emph{Pretraining} & 56 & 69 & 53 & 59 & 59 & 54.8\\
+ CBAM (\emph{Self-Attention})  & 60 & 69 & 56 & 61 & 62 & 56.5 \\
+ $\ell_\infty$ Adversarial Training & 80 & 71 & 72 & 71 & 74 & 71.6 \\
+ Speckle Noise	& 57 & 68 & 60 & 64 & 62 & 56.9\\
+ Style Transfer	& 57 & 68 & 55 & 64 & 61 & 56.7\\
+ AugMix	& 52 & 65 & 46 & 51 & 54 & 54.4\\
+ DeepAugment	& 48 & 60 & 51 & 61 & 55 & 54.2\\
+ DeepAugment+AugMix	& 41 & 53 & 39 & 48 & 45 & 51.7 \\\midrule
ResNet-152 (\emph{Larger Models})  & 67 & 81 & 66 & 74 & 58 & 54.3 \\
\Xhline{2\arrayrulewidth}
\end{tabularx}
\caption{ImageNet-C Blurs (Defocus, Glass, Motion, Zoom) vs Real Blurry Images. All values are error rates and percentages. The rank orderings of the models on Real Blurry Images are similar to the rank orderings for ``ImageNet-C Blur Mean,'' so ImageNet-C's simulated blurs track real-world blur performance.%
}
\label{tab:imagenetblur}
\end{table*}

\noindent \textbf{ImageNet-A.}\quad
ImageNet-A \cite{Hendrycks2019NaturalAE} is an adversarially filtered test set and is constructed based on existing model weaknesses (see \cite{Wang2020I} for another robustness dataset algorithmically determined by model weaknesses). This dataset contains examples that are difficult for a ResNet-50 to classify, so examples solvable by simple spurious cues are are especially infrequent in this dataset.
Results are in \Cref{tab:imageneta}. Notice Res2Net architectures \cite{Gao2019Res2NetAN} can greatly improve accuracy. Results also show that \emph{Larger Models}, \emph{Self-Attention}, and \emph{Pretraining} help, while \emph{Diverse Data Augmentation} usually does not help substantially.

\noindent \textbf{Implications for the Four Methods.}\quad \\
\emph{Larger Models} help with ImageNet-C ($+$), ImageNet-A ($+$), ImageNet-R ($+$), yet does not markedly improve DFR ($-$) performance.\\
\emph{Self-Attention} helps with ImageNet-C ($+$), ImageNet-A ($+$), yet does not help ImageNet-R ($-$) and DFR ($-$) performance.\\
\emph{Diverse Data Augmentation} helps ImageNet-C ($+$), ImageNet-R ($+$), yet does not markedly improve ImageNet-A ($-$), DFR($-$), nor SVSF ($-$) performance.\\
\emph{Pretraining} helps with ImageNet-C ($+$), ImageNet-A ($+$), yet does not markedly improve DFR ($-$) nor ImageNet-R ($-$) performance.

\begin{table*}[h]
\centering
\begin{tabular}{lcccccc}
  Hypothesis       & ImageNet-C & Real Blurry Images & ImageNet-A & ImageNet-R  & DFR & SVSF \\ 
\hline
\emph{Larger Models}    &  $+$ & $+$ & $+$ & $+$ & $-$ \\
\emph{Self-Attention}   & $+$ & $+$ & $+$ & $-$ & $-$  \\
\emph{Diverse Data Augmentation}       & $+$ & $+$ & $-$ & $+$ & $-$ & $-$  \\
\emph{Pretraining}      & $+$ & $+$ & $+$ & $-$ & $-$  \\
\bottomrule
\end{tabular}
\caption{A highly simplified account of each method when tested against different datasets. This table includes ImageNet-A results.}
\label{tab:hypothesissummaryimageneta}
\end{table*}

\begin{table*}[th]
\scriptsize
\begin{center}
{\setlength\tabcolsep{1.1pt}%
\begin{tabular}{@{}l | c | c | c c c | c c c c | c c c  c | c c c c@{}}
 & \multicolumn{2}{c}{} & \multicolumn{3}{c}{Noise} & \multicolumn{4}{c}{Blur} & \multicolumn{4}{c}{Weather} & \multicolumn{4}{c}{Digital}\\
\cline{1-18}
 & \multicolumn{1}{c|}{\,\emph{Clean}\,} & {\,\emph{mCE}\,} & \scriptsize{Gauss.}
    & \scriptsize{Shot} & \scriptsize{Impulse} & \scriptsize{Defocus} & \scriptsize{Glass} & \scriptsize{Motion} & \scriptsize{Zoom} & \scriptsize{Snow} & \scriptsize{Frost} & \scriptsize{Fog} & \scriptsize{Bright} & \scriptsize{Contrast} & \scriptsize{Elastic} & \scriptsize{Pixel} & \scriptsize{JPEG} \\ \hline 
ResNet-50                   & 23.9 & 76.7 & 80 & 82 & 83 & 75 & 89 & 78 & 80 & 78 & 75 & 66 & 57 & 71 & 85 & 77 & 77\\
+ ImageNet-21K \emph{Pretraining}     & 22.4 & 65.8 & 61 & 64 & 63 & 69 & 84 & 68 & 74 & 69 & 71 & 61 & 53 & 53 & 81 & 54 & 63 \\
+ SE (\emph{Self-Attention})  & 22.4 & 68.2 & 63 & 66 & 66 & 71 & 82 & 67 & 74 & 74 & 72 & 64 & 55 & 71 & 73 & 60 & 67 \\
+ CBAM (\emph{Self-Attention})  & 22.4 & 70.0 & 67 & 68 & 68 & 74 & 83 & 71 & 76 & 73 & 72 & 65 & 54 & 70 & 79 & 62 & 67 \\
+ $\ell_\infty$ Adversarial Training      & 46.2 & 94.0 & 91 & 92 & 95 & 97 & 86 & 92 & 88 & 93 & 99 & 118 & 104 & 111 & 90 & 72 & 81  \\
+ Speckle Noise             & 24.2 & 68.3 & 51 & 47 & 55 & 70 & 83 & 77 & 80 & 76 & 71 & 66 & 57 & 70 & 82 & 72 & 69 \\
+ Style Transfer            & 25.4 & 69.3 & 66 & 67 & 68 & 70 & 82 & 69 & 80 & 68 & 71 & 65 & 58 & 66 & 78 & 62 & 70 \\
+ AugMix                    & 22.5 & 65.3 & 67 & 66 & 68 & 64 & 79 & 59 & 64 & 69 & 68 & 65 & 54 & 57 & 74 & 60 & 66  \\
+ DeepAugment               & 23.3 & 60.4 & 49 & 50 & 47 & 59 & 73 & 65 & 76 & 64 & 60 & 58 & 51 & 61 & 76 & 48 & 67  \\
+ DeepAugment + AugMix      & 24.2 & 53.6 & 46 & 45 & 44 & 50 & 64 & 50 & 61& 58 & 57 & 54 & 52 & 48 & 71 & 43 & 61  \\ \midrule
ResNet-152 (\emph{Larger Models}) & 21.7 & 69.3 & 73 & 73 & 76 & 67 & 81 & 66 & 74 & 71 & 68 & 62 & 51 & 67 & 76 & 69 & 65 \\ \midrule
ResNeXt-101 32$\times$\text{8d} (\emph{Larger Models}) & 20.7 & 66.7 & 68 & 69 & 71 & 65 & 79 & 66 & 71 & 69 & 66 & 60 & 50 & 66 & 74 & 61 & 64  \\ 
+ WSL \emph{Pretraining} ($1000\times$ data)           & 17.8 & 51.7 & 49 & 50 & 51 & 53 & 72 & 55 & 63 & 53 & 51 & 42 & 37 & 41 & 67 & 40 & 51 \\
+ DeepAugment + AugMix          & 20.1 & 44.5 & 36 & 35 & 34 & 43 & 55 & 42 & 55 & 48 & 48 & 47 & 43 & 39 & 59 & 34 & 50 \\

\end{tabular}}
\caption{Clean Error, Corruption Error (CE), and mean CE (mCE) values for various models, and training methods on ImageNet-C. The mCE value is computed by averaging across all 15 CE values. A CE value greater than 100 (e.g. adversarial training on contrast) denotes worse performance than AlexNet. DeepAugment+AugMix improves robustness by over 23 mCE.}
\label{tab:imagenetc_table}
\end{center}
\end{table*}

\begin{table*}[t]
\begin{center}
\begin{tabular}{lcc>{\color{gray}}c}
\toprule
                                        &  ImageNet-200 (\%)    & ImageNet-R (\%)   &  Gap      \\ \midrule
ResNet-50 \cite{he2015deep}                              &  7.9                  &  63.9             &  56.0     \\
+ ImageNet-21K \emph{Pretraining} ($10\times$ data)         &  7.0                  &  62.8             &  55.8     \\
+ CBAM (\emph{Self-Attention})                      &  7.0                  &  63.2             &  56.2     \\ 
+ $\ell_\infty$ Adversarial Training       &  25.1                 &  68.6             &  43.5     \\
+ Speckle Noise              &  8.1                  &  62.1             &  54.0     \\
+ Style Transfer             &  8.9                  &  58.5             &  49.6     \\ 
+ AugMix                    &  7.1                  &  58.9             &  51.8     \\
+ DeepAugment                 &  7.5                  &  57.8             &  50.3     \\
+ DeepAugment + AugMix        &  8.0                  &  53.2    &  45.2     \\ \midrule
ResNet-101 (\emph{Larger Models})                             &  7.1                  &  60.7             &  53.6     \\
+ SE (\emph{Self-Attention})                        &  6.7                  &  61.0             &  54.3     \\ \midrule
ResNet-152 (\emph{Larger Models})                             &  6.8                  &  58.7             &  51.9     \\
+ SE (\emph{Self-Attention})                         &  6.6                  &  60.0             &  53.4     \\ \midrule
ResNeXt-101 32$\times$\text{4d} (\emph{Larger Models})        &  6.8                  &  58.0             &  51.2     \\
+ SE (\emph{Self-Attention})    &  5.9                  &  59.6             &  53.7     \\ \midrule
ResNeXt-101 32$\times$\text{8d} (\emph{Larger Models})         &  6.2                  &  57.5             &  51.3     \\
+ WSL \emph{Pretraining} ($1000\times$ data)  &  4.1   &  24.2             &  20.1     \\
+ DeepAugment + AugMix        &  6.1                  &  47.9             &  41.8     \\

\bottomrule
\end{tabular}
\end{center}
\caption{ImageNet-200 and ImageNet-Renditions error rates. ImageNet-21K and WSL Pretraining are \emph{Pretraining} methods, and pretraining gives mixed benefits. CBAM and SE are forms of \emph{Self-Attention}, and these \emph{hurt} robustness. ResNet-152 and ResNeXt-101 32$\times$\text{8d} test the impact of using \emph{Larger Models}, and these help. Other methods augment data, and Style Transfer, AugMix, and DeepAugment provide support for the \emph{Diverse Data Augmentation}.
}
\label{tab:imagenetr_full}
\end{table*}

\begin{table*}[h]
\begin{center}
\begin{tabular}{lc}%
\toprule
                        & ImageNet-A (\%)  \\\midrule%
ResNet-50   & 2.2 \\
+ ImageNet-21K \emph{Pretraining} ($10\times$ data)    & 11.4 \\
+ Squeeze-and-Excitation (\emph{Self-Attention}) & 6.2 \\
+ CBAM (\emph{Self-Attention}) & 6.9 \\
+ $\ell_\infty$ Adversarial Training    & 1.7 \\
+ Style Transfer  & 2.0 \\
+ AugMix  & 3.8\\
+ DeepAugment & 3.5 \\
+ DeepAugment + AugMix  & 3.9\\
\hline
ResNet-152 (\emph{Larger Models})             & 6.1        \\
ResNet-152+Squeeze-and-Excitation (\emph{Self-Attention})           & 9.4        \\
\hline
Res2Net-50 v1b        & 14.6       \\
Res2Net-152 v1b (\emph{Larger Models})       & 22.4        \\
\hline
ResNeXt-101 ($32\times8$d) (\emph{Larger Models})   & 10.2  \\
+ WSL \emph{Pretraining} ($1000\times$ data) & 45.4 \\
+ DeepAugment + AugMix & 11.5 \\
\bottomrule
\end{tabular}
\end{center}
\caption{ImageNet-A top-1 accuracy.}\label{tab:imageneta}
\end{table*}

\section{DeepAugment Details}\label{app:deepaugment}
\paragraph{Pseudocode.} Below is Pythonic pseudocode for DeepAugment. The basic structure of DeepAugment is agnostic to the backbone network used, but specifics such as which layers are chosen for various transforms may vary as the backbone architecture varies. We do not need to train many different image-to-image models to get diverse distortions \cite{zhang2018perceptual,Lee2020NetworkRA}. We only use two existing models, the EDSR super-resolution model \cite{lim2017enhanced} and the CAE image compression model \cite{theis2017lossy}. See full code for such details.

At a high level, DeepAugment processes each image with an image-to-image network. The image-to-image network's weights and feedforward activations are distorted with each pass. The distortion is made possible by, for example, negating the network's weights and applying dropout to the feedforward activations. These modifications were not carefully chosen and demonstrate the utility of mixing together diverse operations without tuning. The resulting image is distorted and saved. This process generates an augmented dataset.

\paragraph{Ablations.} We run ablations on DeepAugment to understand the contributions from the EDSR and CAE models independently. \Cref{tab:imagenetr_deepaug_ablation} contains results of these experiments on ImageNet-R and \Cref{tab:imagenetc_deepaug_ablation} contains results of these experiments on ImageNet-C. In both tables, ``DeepAugment (EDSR)'' and ``DeepAugment (CAE)'' refer to experiments where we only use a single extra augmented training set (+ the standard training set), and train on those images.

\paragraph{Noise2Net.} We show that untrained, randomly sampled neural networks can provide useful deep augmentations, highlighting the efficacy of the DeepAugment approach. While in the main paper we use EDSR and CAE to create DeepAugment augmentations, in this section we explore the use of randomly initialized image-to-image networks to generate diverse image augmentations. We propose a DeepAugment method, \textit{Noise2Net}.

In Noise2Net, the architecture and weights are randomly sampled.
Noise2Net is the composition of several residual blocks: $\text{Block}(x) = x + \varepsilon \cdot f_{\Theta}(x)$, where $\Theta$ is randomly initialized and $\varepsilon$ is a parameter that controls the strength of the augmentation. For all our experiments, we use 4 Res2Net blocks \cite{res2net_gao2019} and $\varepsilon \sim U(0.375, 0.75)$. The weights of Noise2Net are resampled at every minibatch, and the dilation and kernel sizes of all the convolutions used in Noise2Net are randomly sampled every epoch. Hence Noise2Net augments an image to an augmented image by processing the image through a randomly sampled network with random weights.

Recall that in the case of EDSR and CAE, we used networks to generate a static dataset, and then we trained normally on that static dataset. This setup could not be done on-the-fly. That is because we fed in one example at a time with EDSR and CAE. If we pass the entire minibatch through EDSR or CAE, we will end up applying the same augmentation to all images in the minibatch, reducing stochasticity and augmentation diversity. In contrast, Noise2Net enables us to process batches of images on-the-fly and obviates the need for creating a static augmented dataset.

\begin{figure*}[b]
\vspace{-10pt}
\begin{center}
\includegraphics[width=\textwidth]{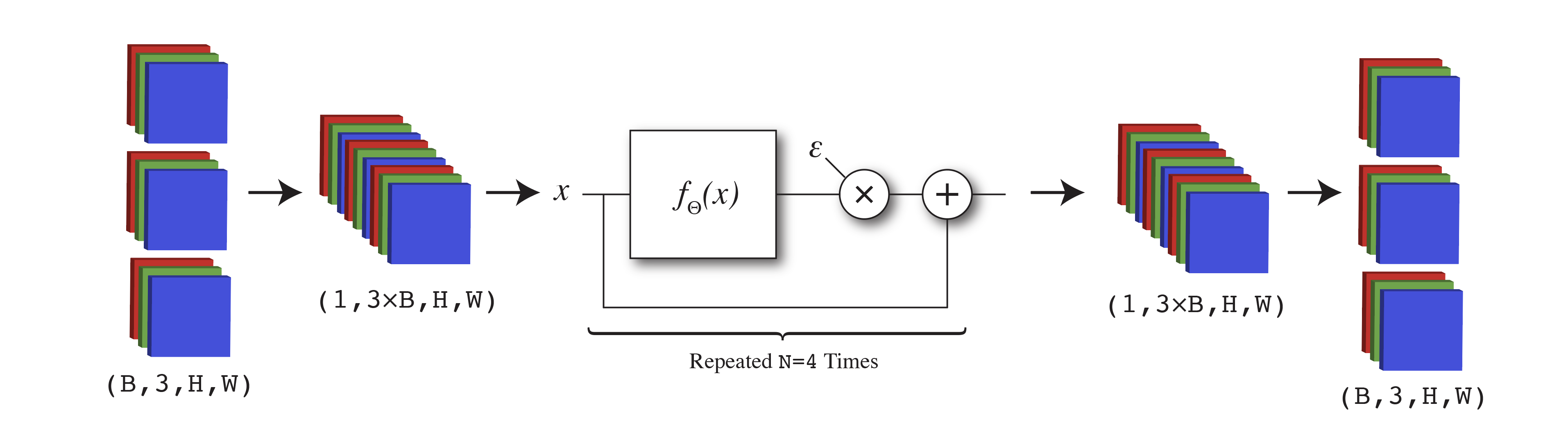}
\end{center}
\caption{
 Parallel augmentation with Noise2Net. We collapse batches to the channel dimension to ensure that different transformations are applied to each image in the batch. Feeding images into the network in the standard way would result in the same augmentation being applied to each image, which is undesirable. The function $f_\Theta(x)$ is a Res2Net block with all convolutions replaced with grouped convolutions.
}\label{fig:noise2net}
\vspace{-10pt}
\end{figure*}

\begin{figure*}[]
\vspace{-10pt}
\begin{center}
\includegraphics[width=\textwidth]{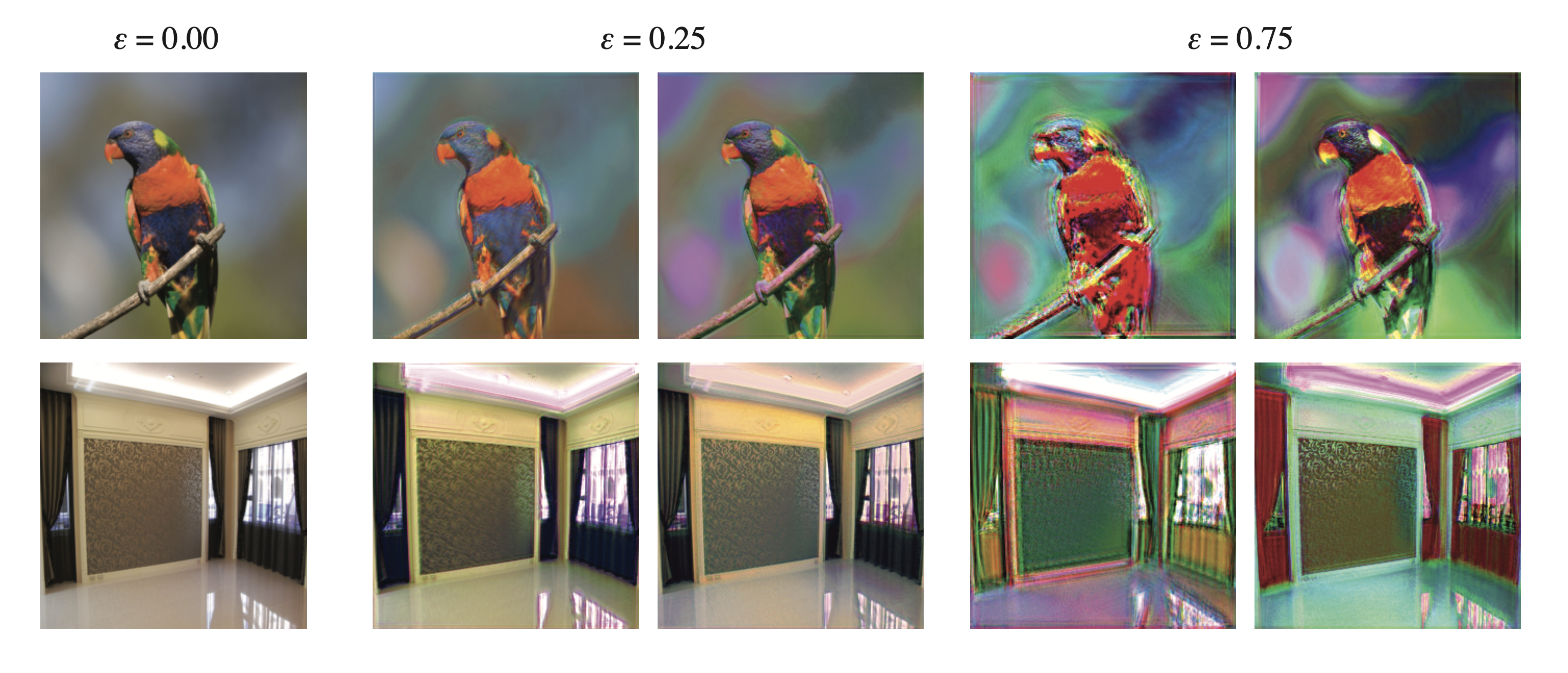}
\end{center}
\caption{
Example outputs of Noise2Net for different values of $\varepsilon$. Note $\varepsilon = 0$ is the original image.
}\label{fig:noise2net_examples}
\vspace{-10pt}
\end{figure*}

In Noise2Net, each example is processed differently in parallel, so we generate more diverse augmentations in real-time. To make this possible, we use grouped convolutions. A grouped convolution with number of groups = $N$ will take a set of $kN$ channels as input, and apply $N$ independent convolutions on channels $\{1, \ldots, k\}, \{k + 1, \ldots, 2k\}, \ldots, \{(N-1)k + 1, \ldots, Nk\}$. Given a minibatch of size $B$, we can apply a randomly initialized grouped convolution with $N=B$ groups in order to apply a different random convolutional filter to each element in the batch in a single forward pass. By replacing all the convolutions in each Res2Net block with a grouped convolution and randomly initializing network weights, we arrive at Noise2Net, a variant of DeepAugment. See \Cref{fig:noise2net} for a high-level overview of Noise2Net and \Cref{fig:noise2net_examples} for sample outputs.

We evaluate the Noise2Net variant of DeepAugment on ImageNet-R. \Cref{tab:imagenetr_deepaug_ablation} shows that it outperforms the EDSR and CAE variants of DeepAugment, even though the network architecture is randomly sampled, its weights are random, and the network is not trained. This demonstrates the flexibility of the DeepAugment approach. Below is Pythonic pseudocode for training a classifier using the Noise2Net variant of DeepAugment.

\begin{lstlisting}[language=Python]
def main():
    net.apply_weights(deepAugment_getNetwork())  # EDSR, CAE, ...
    for image in dataset:  # May be the ImageNet training set
        if np.random.uniform() < 0.05:  # Arbitrary refresh prob
            net.apply_weights(deepAugment_getNetwork())
        new_image = net.deepAugment_forwardPass(image)

def deepAugment_getNetwork():
    weights = load_clean_weights()
    weight_distortions = sample_weight_distortions()
    for d in weight_distortions:
        weights = apply_distortion(d, weights)
    return weights

def sample_weight_distortions():
    distortions = [
        negate_weights,
        zero_weights,
        flip_transpose_weights,
        ...
    ]
    
    return random_subset(distortions)
    
def sample_signal_distortions():
    distortions = [
        gelu,
        negate_signal_random_mask,
        flip_signal,        
        ...
    ]
    
    return random_subset(distortions)


class Network():
    def apply_weights(weights):
        ... # Apply given weight tensors to network
    
    # Clean forward pass. Compare to deepAugment_forwardPass()
    def clean_forwardPass(X):
        X = network.block1(X)
        X = network.block2(X)
        ...
        X = network.blockN(X)
        return X

    # Our forward pass. Compare to clean_forwardPass()
    def deepAugment_forwardPass(X):
        # Returns a list of distortions, each of which 
        # will be applied at a different layer.
        signal_distortions = sample_signal_distortions()
	       
        X = network.block1(X)
        apply_layer_1_distortions(X, signal_distortions)
        X = network.block2(X)
        apply_layer_2_distortions(X, signal_distortions)
        ...
        apply_layer_N-1_distortions(X, signal_distortions)
        X = network.blockN(X)
        apply_layer_N_distortions(X, signal_distortions)
	
    return X
\end{lstlisting}

\begin{lstlisting}[language=Python]
def train_one_epoch(classifier, batch_size, dataloader):
    noise2net = Noise2Net(batch_size=batch_size)
    for batch, target in dataloader: 
        noise2net.reload_weights()
        noise2net.set_epsilon(random.uniform(0.375, 0.75))
        logits = model(noise2net.forward(batch))
        ... # Calculate loss and backrop

def train():
    for epoch in range(epochs):
        train_one_epoch(classifier, batch_size, dataloader)

class DeepAugment_Noise2Net:
    def __init__(self, batch_size=5):
        self.block1 = Res2NetBlock(batch_size=batch_size)
        self.block2 = Res2NetBlock(batch_size=batch_size)
        self.block3 = Res2NetBlock(batch_size=batch_size)
        self.block4 = Res2NetBlock(batch_size=batch_size)

    def reload_weights(self):
        ... # Reload Network parameters

    def set_epsilon(self, new_eps):
        self.epsilon = new_eps

    def forward(self, x):
        x = x + self.block1(x) * self.epsilon
        x = x + self.block2(x) * self.epsilon
        x = x + self.block3(x) * self.epsilon
        x = x + self.block4(x) * self.epsilon
        return x
\end{lstlisting}

\begin{table*}[]
\scriptsize
\begin{center}
{\setlength\tabcolsep{0.9pt}%
\begin{tabular}{@{}l  c | c | c c c | c c c c | c c c  c | c c c c@{}}
 & \multicolumn{2}{c}{} & \multicolumn{3}{c}{Noise} & \multicolumn{4}{c}{Blur} & \multicolumn{4}{c}{Weather} & \multicolumn{4}{c}{Digital}\\
\cline{1-18}
 & \multicolumn{1}{c|}{\,\emph{Clean}\,} & {\,\emph{mCE}\,} & \scriptsize{Gauss.}
    & \scriptsize{Shot} & \scriptsize{Impulse} & \scriptsize{Defocus} & \scriptsize{Glass} & \scriptsize{Motion} & \scriptsize{Zoom} & \scriptsize{Snow} & \scriptsize{Frost} & \scriptsize{Fog} & \scriptsize{Bright} & \scriptsize{Contrast} & \scriptsize{Elastic} & \scriptsize{Pixel} & \scriptsize{JPEG} \\ \hline 
ResNet-50                       & 23.9 & 76.7 & 80 & 82 & 83 & 75 & 89 & 78 & 80 & 78 & 75 & 66 & 57 & 71 & 85 & 77 & 77 \\
+ DeepAugment (EDSR)            & 23.5 & 64.0 & 56 & 57 & 54 & 64 & 77 & 71 & 78 & 68 & 64 & 64 & 55 & 64 & 78 & 46 & 67 \\
+ DeepAugment (CAE)             & 23.2 & 67.0 & 58 & 60 & 62 & 62 & 75 & 73 & 77 & 68 & 66 & 60 & 52 & 66 & 80 & 63 & 78 \\
+ DeepAugment (Both)            & 23.3 & 60.4 & 49 & 50 & 47 & 59 & 73 & 65 & 76 & 64 & 60 & 58 & 51 & 61 & 76 & 48 & 67 \\

\end{tabular}}
\caption{Clean Error, Corruption Error (CE), and mean CE (mCE) values for DeepAugment ablations on ImageNet-C. The mCE value is computed by averaging across all 15 CE values.}
\label{tab:imagenetc_deepaug_ablation}
\end{center}
\end{table*}

\newpage

\begin{table*}[]
\begin{center}
\begin{tabular}{lcc>{\color{gray}}c}
\toprule
                                    &  ImageNet-200 (\%)    & ImageNet-R (\%)   &  Gap      \\ \midrule
ResNet-50                           &  7.9                  &  63.9             &  56.0     \\
+ DeepAugment (EDSR)                &  7.9                  &  60.3             &  55.1    \\
+ DeepAugment (CAE)                 &  7.6                  &  58.5             &  50.9    \\
+ DeepAugment (EDSR + CAE)          &  7.5                  &  57.8             &  50.3     \\
+ DeepAugment (Noise2Net)           &  7.2                  &  57.6             &  50.4     \\ \midrule
+ DeepAugment (All 3)               &  7.4                  &  56.0             &  48.6     \\

\bottomrule
\end{tabular}
\end{center}
\caption{DeepAugment ablations on ImageNet-200 and ImageNet-Renditions.}
\label{tab:imagenetr_deepaug_ablation}
\end{table*}

\newpage
\section{Further Dataset Descriptions}
\paragraph{ImageNet-R Classes.}\label{app:classes}
The 200 ImageNet classes and their WordNet IDs in ImageNet-R are as follows.

Goldfish, \quad great white shark, \quad hammerhead, \quad stingray, \quad hen, \quad ostrich, \quad goldfinch, \quad junco, \quad bald eagle, \quad vulture, \quad newt, \quad axolotl, \quad tree frog, \quad iguana, \quad African chameleon, \quad cobra, \quad scorpion, \quad tarantula, \quad centipede, \quad peacock, \quad lorikeet, \quad hummingbird, \quad toucan, \quad duck, \quad goose, \quad black swan, \quad koala, \quad jellyfish, \quad snail, \quad lobster, \quad hermit crab, \quad flamingo, \quad american egret, \quad pelican, \quad king penguin, \quad grey whale, \quad killer whale, \quad sea lion, \quad chihuahua, \quad shih tzu, \quad afghan hound, \quad basset hound, \quad beagle, \quad bloodhound, \quad italian greyhound, \quad whippet, \quad weimaraner, \quad yorkshire terrier, \quad boston terrier, \quad scottish terrier, \quad west highland white terrier, \quad golden retriever, \quad labrador retriever, \quad cocker spaniels, \quad collie, \quad border collie, \quad rottweiler, \quad german shepherd dog, \quad boxer, \quad french bulldog, \quad saint bernard, \quad husky, \quad dalmatian, \quad pug, \quad pomeranian, \quad chow chow, \quad pembroke welsh corgi, \quad toy poodle, \quad standard poodle, \quad timber wolf, \quad hyena, \quad red fox, \quad tabby cat, \quad leopard, \quad snow leopard, \quad lion, \quad tiger, \quad cheetah, \quad polar bear, \quad meerkat, \quad ladybug, \quad fly, \quad bee, \quad ant, \quad grasshopper, \quad cockroach, \quad mantis, \quad dragonfly, \quad monarch butterfly, \quad starfish, \quad wood rabbit, \quad porcupine, \quad fox squirrel, \quad beaver, \quad guinea pig, \quad zebra, \quad pig, \quad hippopotamus, \quad bison, \quad gazelle, \quad llama, \quad skunk, \quad badger, \quad orangutan, \quad gorilla, \quad chimpanzee, \quad gibbon, \quad baboon, \quad panda, \quad eel, \quad clown fish, \quad puffer fish, \quad accordion, \quad ambulance, \quad assault rifle, \quad backpack, \quad barn, \quad wheelbarrow, \quad basketball, \quad bathtub, \quad lighthouse, \quad beer glass, \quad binoculars, \quad birdhouse, \quad bow tie, \quad broom, \quad bucket, \quad cauldron, \quad candle, \quad cannon, \quad canoe, \quad carousel, \quad castle, \quad mobile phone, \quad cowboy hat, \quad electric guitar, \quad fire engine, \quad flute, \quad gasmask, \quad grand piano, \quad guillotine, \quad hammer, \quad harmonica, \quad harp, \quad hatchet, \quad jeep, \quad joystick, \quad lab coat, \quad lawn mower, \quad lipstick, \quad mailbox, \quad missile, \quad mitten, \quad parachute, \quad pickup truck, \quad pirate ship, \quad revolver, \quad rugby ball, \quad sandal, \quad saxophone, \quad school bus, \quad schooner, \quad shield, \quad soccer ball, \quad space shuttle, \quad spider web, \quad steam locomotive, \quad scarf, \quad submarine, \quad tank, \quad tennis ball, \quad tractor, \quad trombone, \quad vase, \quad violin, \quad military aircraft, \quad wine bottle, \quad ice cream, \quad bagel, \quad pretzel, \quad cheeseburger, \quad hotdog, \quad cabbage, \quad broccoli, \quad cucumber, \quad bell pepper, \quad mushroom, \quad Granny Smith, \quad strawberry, \quad lemon, \quad pineapple, \quad banana, \quad pomegranate, \quad pizza, \quad burrito, \quad espresso, \quad volcano, \quad baseball player, \quad scuba diver, \quad acorn.

n01443537, \quad n01484850, \quad n01494475, \quad n01498041, \quad n01514859, \quad n01518878, \quad n01531178, \quad n01534433, \quad n01614925, \quad n01616318, \quad n01630670, \quad n01632777, \quad n01644373, \quad n01677366, \quad n01694178, \quad n01748264, \quad n01770393, \quad n01774750, \quad n01784675, \quad n01806143, \quad n01820546, \quad n01833805, \quad n01843383, \quad n01847000, \quad n01855672, \quad n01860187, \quad n01882714, \quad n01910747, \quad n01944390, \quad n01983481, \quad n01986214, \quad n02007558, \quad n02009912, \quad n02051845, \quad n02056570, \quad n02066245, \quad n02071294, \quad n02077923, \quad n02085620, \quad n02086240, \quad n02088094, \quad n02088238, \quad n02088364, \quad n02088466, \quad n02091032, \quad n02091134, \quad n02092339, \quad n02094433, \quad n02096585, \quad n02097298, \quad n02098286, \quad n02099601, \quad n02099712, \quad n02102318, \quad n02106030, \quad n02106166, \quad n02106550, \quad n02106662, \quad n02108089, \quad n02108915, \quad n02109525, \quad n02110185, \quad n02110341, \quad n02110958, \quad n02112018, \quad n02112137, \quad n02113023, \quad n02113624, \quad n02113799, \quad n02114367, \quad n02117135, \quad n02119022, \quad n02123045, \quad n02128385, \quad n02128757, \quad n02129165, \quad n02129604, \quad n02130308, \quad n02134084, \quad n02138441, \quad n02165456, \quad n02190166, \quad n02206856, \quad n02219486, \quad n02226429, \quad n02233338, \quad n02236044, \quad n02268443, \quad n02279972, \quad n02317335, \quad n02325366, \quad n02346627, \quad n02356798, \quad n02363005, \quad n02364673, \quad n02391049, \quad n02395406, \quad n02398521, \quad n02410509, \quad n02423022, \quad n02437616, \quad n02445715, \quad n02447366, \quad n02480495, \quad n02480855, \quad n02481823, \quad n02483362, \quad n02486410, \quad n02510455, \quad n02526121, \quad n02607072, \quad n02655020, \quad n02672831, \quad n02701002, \quad n02749479, \quad n02769748, \quad n02793495, \quad n02797295, \quad n02802426, \quad n02808440, \quad n02814860, \quad n02823750, \quad n02841315, \quad n02843684, \quad n02883205, \quad n02906734, \quad n02909870, \quad n02939185, \quad n02948072, \quad n02950826, \quad n02951358, \quad n02966193, \quad n02980441, \quad n02992529, \quad n03124170, \quad n03272010, \quad n03345487, \quad n03372029, \quad n03424325, \quad n03452741, \quad n03467068, \quad n03481172, \quad n03494278, \quad n03495258, \quad n03498962, \quad n03594945, \quad n03602883, \quad n03630383, \quad n03649909, \quad n03676483, \quad n03710193, \quad n03773504, \quad n03775071, \quad n03888257, \quad n03930630, \quad n03947888, \quad n04086273, \quad n04118538, \quad n04133789, \quad n04141076, \quad n04146614, \quad n04147183, \quad n04192698, \quad n04254680, \quad n04266014, \quad n04275548, \quad n04310018, \quad n04325704, \quad n04347754, \quad n04389033, \quad n04409515, \quad n04465501, \quad n04487394, \quad n04522168, \quad n04536866, \quad n04552348, \quad n04591713, \quad n07614500, \quad n07693725, \quad n07695742, \quad n07697313, \quad n07697537, \quad n07714571, \quad n07714990, \quad n07718472, \quad n07720875, \quad n07734744, \quad n07742313, \quad n07745940, \quad n07749582, \quad n07753275, \quad n07753592, \quad n07768694, \quad n07873807, \quad n07880968, \quad n07920052, \quad n09472597, \quad n09835506, \quad n10565667, \quad n12267677.

\paragraph{Street View StoreFronts.} The classes are
\begin{multicols}{3}
    \begin{itemize}
        \item auto shop
        \item bakery
        \item bank
        \item beauty salon
        \item car dealer
        \item car wash
        \item cell phone store
        \item dentist
        \item discount store
        \item dry cleaner
        \item furniture store
        \item gas station
        \item gym
        \item hardware store
        \item hotel
        \item liquor store
        \item pharmacy
        \item religious institution
        \item storage facility
        \item veterinary care.
    \end{itemize}
\end{multicols}

\paragraph{DeepFashion Remixed.} The classes are
\begin{multicols}{3}
    \begin{itemize}
        \item short sleeve top
        \item long sleeve top
        \item short sleeve outerwear
        \item long sleeve outerwear
        \item vest
        \item sling
        \item shorts
        \item trousers
        \item skirt
        \item short sleeve dress
        \item long sleep dress
        \item vest dress
        \item sling dress.
    \end{itemize}
\end{multicols}

Size (small, moderate, or large) defines how much of the image the article of clothing takes up. Occlusion (slight, medium, or heavy) defines the degree to which the object is occluded from the camera. Viewpoint (front, side/back, or not worn) defines the camera position relative to the article of clothing. Zoom (no zoom, medium, or large) defines how much camera zoom was used to take the picture.  

\begin{table*}[t]
\begin{center}
\begin{tabular}{lc}%
\toprule
                        & Represented Distribution Shifts           \\\midrule
ImageNet-Renditions     & artistic renditions (cartoons, graffiti, embroidery, graphics, origami, \\
                        & paintings, sculptures, sketches, tattoos, toys, ...)       \\
DeepFashion Remixed     & occlusion, size, viewpoint, zoom         \\
StreetView StoreFronts  & camera, capture year, country             \\
\bottomrule
\end{tabular}
\caption{Various distribution shifts represented in our three new benchmarks. ImageNet-Renditions is a new test set for ImageNet trained models measuring robustness to various object renditions. DeepFashion Remixed and StreetView StoreFronts each contain a training set and multiple test sets capturing a variety of distribution shifts.}
\label{tab:datasetshifts}
\end{center}
\end{table*}

\begin{table*}[]
\begin{center}
\begin{tabular}{lcc}%
\toprule
                        & Training set              &  Testing images \\\midrule
ImageNet-R     & 1281167 &  30000          \\
DFR     & 48000                    &  42640, 7440, 28160, 10360, 480, 11040, 10520, 10640    \\
SVSF  & 200000                   &  10000, 10000, 10000, 8195, 9788  \\
\bottomrule
\end{tabular}
\caption{Number of images in each training and test set. ImageNet-R training set refers to the ILSVRC 2012 training set \cite{ImageNet}. DeepFashion Remixed test sets are: in-distribution, occlusion - none/slight, occlusion - heavy, size - small, size - large, viewpoint - frontal, viewpoint - not-worn, zoom-in - medium, zoom-in - large. StreetView StoreFronts test sets are: in-distribution, capture year - 2018, capture year - 2017, camera system - new, country - France.}
\label{tab:datasetsizes}
\end{center}
\end{table*}

\end{appendices}

\end{document}